\documentclass[aps,prx,twocolumn,superscriptaddress,nofootinbib,longbibliography]{revtex4-2}

\usepackage[T1]{fontenc}
\usepackage{lmodern}
\usepackage{amsmath,amssymb,bm,mathtools}
\usepackage{graphicx}
\usepackage{array}[=2016-10-06]
\usepackage{booktabs}
\usepackage{siunitx}
\usepackage{xcolor}
\usepackage{microtype}
\usepackage{tikz}
\usetikzlibrary{arrows.meta,positioning,calc,fit}
\usepackage{placeins}
\usepackage{float}
\usepackage[colorlinks=true,allcolors=blue!55!black]{hyperref}

\newcommand{\NRMSE}{\mathrm{NRMSE}}
\newcommand{\J}{\mathcal{J}}

\newcommand{\suppsection}[2]{%
  \refstepcounter{section}%
  \section*{SUPP.~\thesection.\ #1}%
  \label{#2}%
}

\begin{document}

\title{Delayed Optimizer-State Transport Shapes Short-Horizon Training Decisions}

\author{Jinhui Guo}
\email{guojh23@buaa.edu.cn}
\affiliation{School of Physics, Beihang University, Beijing 100191, China}

\begin{abstract}
Adaptive optimizers retain gradient history in moment variables, allowing a local change in loss weighting to alter later updates. We examine whether this delayed transport is large enough to change prospective short-horizon decisions. On committed future-minibatch sequences, we differentiate eight-step AdamW trajectories through the complete model--optimizer state and select exposure-matched Math--Code loss schedules before independent evaluation. Across 12 unused 0.3M Transformer histories, full transport lowers token-disjoint loss relative to an optimizer-aware immediate derivative in 10/12 histories (mean benefit $4.71\times10^{-4}$; exact one-sided sign test, $p=0.0193$). The two controllers act equally often but select different schedules in 60/96 windows. Crossed checkpoint--future-path tests attribute this reordering to the interaction between optimizer state and near-future data, while an independent Ising--CNN experiment shows that deleting moment-state transport destroys accurate response prediction. Full-transport scores also concentrate exact-rollout winners in larger candidate libraries, focusing finite-amplitude evaluation on a shortlist. On these committed short paths, optimizer memory and near-future data order are therefore actionable components of the training state, providing a mechanism-based criterion for when finite-horizon rather than one-step intervention is required.
\end{abstract}

\maketitle

\section{Introduction}

Adaptive optimizers endow neural network training with internal dynamical state and memory. A change in data or task weighting alters both the immediate parameter update and the optimizer moments that shape later updates~\cite{kingma2014adam,loshchilov2019adamw}. A local intervention can therefore propagate along the subsequent training trajectory: interventions that look similar immediately may rank differently under an objective evaluated several steps later. Choosing when to intervene is consequently a dynamical question rather than a purely local one.

Training-data composition is an important controllable component of modern learning systems. Data-mixture and curriculum methods optimize global or time-dependent domain exposure using proxy losses, current losses, rewards, or gradients~\cite{albalak2024survey,liu2025mixturesurvey,xie2023doremi,fan2024doge,liu2024regmix,chen2023skillit,albalak2023odm,fan2024dga,li2025pike}. Unrolled and bilevel differentiation relate downstream objectives to earlier training choices~\cite{maclaurin2015reversible,franceschi2018bilevel}, while influence and trajectory methods quantify how examples or updates affect later predictions~\cite{koh2017,pruthi2020,chen2021hydra,park2023trak}. Domain order can produce noncommuting updates, with optimizer memory promoting ordering effects to leading order~\cite{rukhovich2025commute,sweeney2026optmemory}; optimizer-aware attribution can likewise differ from parameter-gradient attribution~\cite{ding2026adamshapley}. We resolve the response to a short intervention into injection, subsequent model--optimizer transport, and later readout, and quantify how transport changes the ranking of finite actions.

Nonequilibrium response theory provides useful language for this problem. In a driven system, a perturbation propagates through internal degrees of freedom before affecting a later observable~\cite{kubo1957,ruelle1998response}; eliminating those degrees of freedom can introduce memory into the reduced dynamics~\cite{zwanzig1961memory,mori1965transport}. AdamW has the same structural feature: its moment buffers retain past-gradient information, leaving parameter-only dynamics with hidden-history dependence. Differentiating the realized optimization trajectory traces the perturbation from its source to a later objective, without assuming equilibrium or fluctuation--dissipation.

Let $x_t$ denote the augmented training state required for a locally Markovian update, including the model parameters and optimizer variables. For a control perturbation applied at step $s$, the linear response of an observable $y_t$ evaluated at a later step $t$ is
\begin{equation}
    \chi_y(t,s)=C_{y,t}\Phi(t,s+1)B_s,
    \label{eq:chi-intro}
\end{equation}
where $B_s$ injects the intervention into the training state, $\Phi(t,s+1)$ is the ordered product of intervening update Jacobians, and $C_{y,t}$ projects the propagated perturbation onto the observable. Equation~\eqref{eq:chi-intro} is the standard unrolled hypergradient in source--transport--readout form. Here $B_s$ is set by a controlled domain-loss contrast, $\Phi$ transports parameter and optimizer perturbations, and $C_{y,t}$ measures a later held-out objective. Deleting selected optimizer-coordinate tangents while keeping the nominal trajectory fixed tests how those coordinates contribute to transmission of the finite-time response.

We study this question in a paired two-domain training setup, using exposure-matched, eight-step Math--Code loss schedules in a byte-level Transformer. Both domain minibatches are evaluated at every step, fixing sample allocation so that interventions differ only in the temporal placement of their loss weights. Actions are selected on committed future-minibatch paths and locked before an independent token-disjoint readout. A crossed checkpoint--future-path experiment separates the current optimizer state from the subsequent training drive. An Ising--CNN benchmark tests the same transport mechanism under a different architecture, data-generating process, and scientific readout. For larger schedule libraries, finite-time scores rank candidates before exact finite-amplitude rollouts choose among a shortlist.

\begin{figure*}[t]
  \centering
  \resizebox{0.98\textwidth}{!}{\begin{tikzpicture}[
  font=\small\sffamily,
  >=Latex,
  node distance=6mm and 9mm,
  box/.style={draw, rounded corners=1.3mm, minimum height=12mm, align=center, inner sep=2.8mm, line width=0.7pt},
  state/.style={box, fill=blue!7, draw=blue!55!black},
  source/.style={box, fill=orange!10, draw=orange!75!black},
  readout/.style={box, fill=green!9, draw=green!55!black},
  decision/.style={box, fill=purple!7, draw=purple!55!black},
  note/.style={align=center, font=\scriptsize\sffamily},
  arr/.style={->, line width=0.85pt},
  dashedarr/.style={->, dashed, line width=0.75pt, gray!80}
]
  \node[source] (mix) {mixture perturbation\\$p_s=p_0+\delta p_s$};
  \node[source, right=of mix] (B) {source tangent\\$B_s=\partial_pF_s$};
  \node[state, right=of B, minimum width=29mm] (x1) {full state\\$x_{s+1}=(\theta,m_{\rm Adam},v_{\rm Adam},k)_{s+1}$};
  \node[state, right=of x1, minimum width=26mm] (phi) {pathwise propagator\\$\Phi(t,s+1)=A_{t-1}\cdots A_{s+1}$};
  \node[readout, right=of phi] (C) {observable readout\\$C_{y,t}=(\nabla_x y_t)^{\mathsf T}$};
  \node[readout, right=of C] (dy) {response prediction\\$\delta y_t=\sum_{s<t}\chi_y(t,s)\delta p_s$};

  \draw[arr, orange!80!black] (mix) -- (B);
  \draw[arr] (B) -- node[above,note] {inject} (x1);
  \draw[arr] (x1) -- node[above,note] {transport} (phi);
  \draw[arr] (phi) -- node[above,note] {project} (C);
  \draw[arr, green!50!black] (C) -- (dy);

  \node[note, below=3mm of phi] (chi) {$\displaystyle \chi_y(t,s)=C_{y,t}\,\Phi(t,s+1)\,B_s$\\pathwise finite-time response};

  \node[state, below=18mm of x1, fill=red!6, draw=red!65!black] (ablate) {memory intervention\\reset $\delta m_{\rm Adam},\delta v_{\rm Adam}$ after each step};
  \draw[dashedarr, red!70!black] (B.south) |- (ablate.west);
  \draw[dashedarr, red!70!black] (ablate.east) -| node[pos=0.22,below,note] {tangent-path deletion} (phi.south);

  \node[decision, below=18mm of C, minimum width=31mm] (cert) {discrete branch screen\\sampled amplitude homotopy\\same clipping branch};
  \node[decision, right=of cert, minimum width=31mm] (choose) {lock action before test\\$u^*=\arg\min_{u\in\mathcal U_{\rm branch}}\widehat J(u)$};
  \draw[arr, purple!70!black] (dy.south) -- ++(0,-6mm) -| (choose.north);
  \draw[arr, purple!70!black] (cert) -- (choose);
  \coordinate (budgetroute) at ([yshift=-7mm]ablate.south);
  \draw[arr, purple!70!black] (mix.south) |- (budgetroute) -| ([xshift=-5mm]cert.west) -- (cert.west);
  \node[note, anchor=north, text width=43mm] at ([xshift=-8mm,yshift=-0.5mm]budgetroute)
    {fixed aggregate coefficient; $\sum_s\delta p_s=0$ only for order-only tests};

  \node[anchor=west, font=\bfseries\sffamily] at ([yshift=6mm]mix.north west) {A. Mechanism};
  \node[anchor=west, font=\bfseries\sffamily] at ([yshift=6mm]ablate.north west) {B. Ablation and prospective decision};
\end{tikzpicture}}
  \caption{\textbf{Pathwise training response separates source, transport, and readout.} A local loss-weight perturbation enters the augmented model--optimizer state through $B_s$, propagates through the finite-time Jacobian product $\Phi$, and is projected by the observable-specific readout $C_{y,t}$. The same decomposition supports the three tests used below: deleting moment-state tangent channels, comparing full transport with an optimizer-aware immediate response, and ranking finite temporal schedules before exact reranking in large candidate libraries. Actions and terminal states are locked before the independent readout.}
  \label{fig:framework}
\end{figure*}
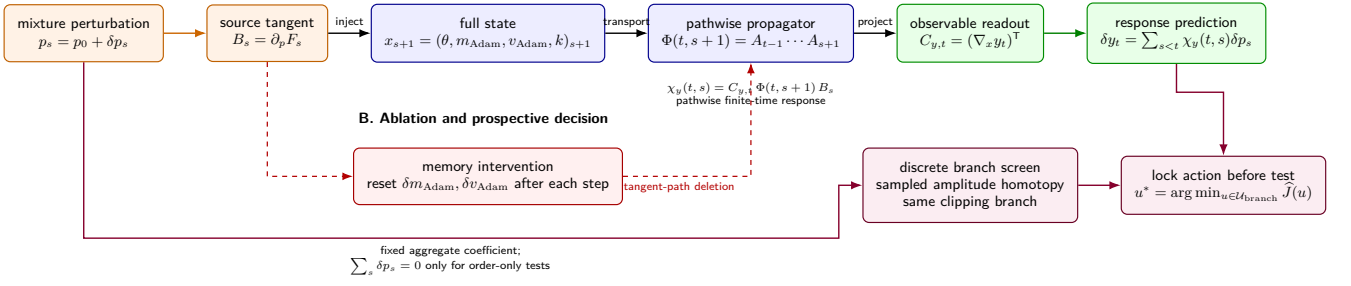

We find that delayed optimizer-state transport materially reorders finite interventions and improves held-out outcomes under prospectively locked decisions. Optimizer memory thereby becomes an actionable component of the training state, with consequences for finite schedule selection. For short committed paths, the results motivate a concrete design principle: candidate schedules should be evaluated using both the augmented model--optimizer state and the near-future data path, while transport-based scores can focus exact finite-amplitude evaluation on a small shortlist. Observing the same mechanism in a Transformer language-modeling system and an independently anchored Ising--CNN benchmark supports its relevance beyond a single architecture, although its range at larger scales and under unknown future paths remains to be established. Figure~\ref{fig:framework} summarizes the dynamical decomposition and experimental design.

\section{Finite-time response and actions}
\label{sec:response-theory}

This section connects a local loss-weight perturbation to a finite schedule decision. The full-state tangent defines the source--transport--readout response along a fixed future path; coordinate deletion then isolates transmission through optimizer moments. Contracting the resulting source-time derivatives with exposure-matched schedules gives the local scores used for finite action selection.

\subsection{Full optimizer-state response}
\label{sec:fullstate}

Consider a fixed realization of the training path. Let $x_t=(\theta_t,z_t^{\mathrm{opt}})$ denote the augmented state needed to make a single optimizer step locally Markovian. For AdamW, $z_t^{\mathrm{opt}}$ contains the first and second moments and the optimizer clock. The update can be written as
\begin{equation}
    x_{t+1}=F_t(x_t,p_t,\xi_t),
    \label{eq:update}
\end{equation}
where $p_t$ controls the relative weight of two paired domains and $\xi_t$ specifies the realized minibatches and any remaining stochasticity. Here $F_t$ denotes the complete one-step AdamW update map, including the moment updates, bias correction, weight decay, clipping branch, and parameter update. Its explicit $t$-dependence accommodates the optimizer clock and any prescribed learning-rate schedule.
With the minibatch pair fixed, the training loss is
\begin{equation}
    \ell_t(\theta,p)=p\,\ell_{A,t}(\theta)+(1-p)\,\ell_{B,t}(\theta),
\end{equation}
Here $\theta$ denotes the model parameters, while $\ell_{A,t}$ and $\ell_{B,t}$ are the losses on the paired domain-$A$ and domain-$B$ minibatches presented at step $t$. The symbol $p$ is the scalar loss-weight argument, and $p_t$ is its realized value at step $t$. Because both minibatches are evaluated at every step, $p_t$ changes their relative loss weights without changing sample allocation.

Linearization about a nominal trajectory, within a fixed differentiable branch of the update map, gives
\begin{equation}
    \delta x_{t+1}=A_t\delta x_t+B_t\delta p_t,
    \qquad
    A_t=\frac{\partial F_t}{\partial x_t},\quad
    B_t=\frac{\partial F_t}{\partial p_t},
    \label{eq:tangent}
\end{equation}
where $\delta x_t$ and $\delta p_t$ denote first-order variations and all derivatives are evaluated on the nominal path. For a finite displacement, the omitted remainder is $O(\|(\delta x_t,\delta p_t)\|^2)$ provided that the perturbation stays on the same smooth branch. For a scalar observable $y_t=h_t(x_t)$, its readout covector is
\begin{equation}
    C_{y,t}=\left(\nabla_x h_t(x_t)\right)^{\mathsf T}.
\end{equation}
For $\delta x_{t_0}=0$, the response of the linearized system at time $t$ is
\begin{equation}\label{eq:response}
\begin{aligned}
    &\delta y_t=\sum_{s=t_0}^{t-1} C_{y,t}\Phi(t,s+1)B_s\,\delta p_s,  \\
    &\Phi(t,s+1)=A_{t-1}\cdots A_{s+1},
\end{aligned}
\end{equation}
with $\Phi(s+1,s+1)=I$. In this factorization, $B_s\delta p_s$ is the perturbation injected into the augmented state, $\Phi(t,s+1)$ is its transport through the intervening updates, and $C_{y,t}$ is the final projection onto the observable.

For a terminal objective $\J_T=\J_T(x_T)$ that depends on earlier controls only through $x_T$, define
\begin{equation}
    C_T=\left(\nabla_x\J_T(x_T)\right)^{\mathsf T}.
\end{equation}
The derivative with respect to a control applied at step $s$ is then
\begin{equation}
    \frac{\partial \J_T}{\partial p_s}
    =C_T\Phi(T,s+1)B_s.
    \label{eq:hypergradient-equivalence}
\end{equation}
Equation~\eqref{eq:hypergradient-equivalence} is the standard unrolled hypergradient in source--transport--readout form. The explicit AdamW tangent and its coordinate-resolved closure checks are given in Supps.~\ref{sup:sub-adamw-tangent} and \ref{sup:sub-coordinate-transport}, respectively.

The ordered product $\Phi$ depends on the source time and on the realized future path. Moving a signed reweighting within a control window changes the Jacobians that follow it and hence its terminal effect. Each $A_t$ is set by the evolving model--optimizer state and the minibatch at that step, so a different ordering of future updates generally produces a different response. The exposure-matched schedules below probe this source-time dependence.

Let $L_A$ and $L_B$ be terminal losses evaluated on paired domain-$A$ and domain-$B$ readout data. Their antisymmetric and symmetric combinations are
\begin{equation}
    m=\frac{L_A-L_B}{2},
    \qquad
    e=\frac{L_A+L_B}{2}.
\end{equation}
The readout $m$ measures redistribution between the domains, and $e$ measures the common-mode loss. The underlying state perturbation is the same; only the terminal projection differs.

\subsection{Optimizer memory as a transport channel}

To further identify which state coordinates transmit the response, we selectively remove perturbations in the Adam moment coordinates while retaining the same nominal trajectory and future minibatches. The resulting \emph{memory-deleted} tangent isolates their contribution to the transport factor $\Phi$ and provides a coordinate-level diagnostic of moment-mediated transmission in the Adam state representation.

The origin of this contribution is already visible in the first-moment equation. Let $r_s=\nabla_\theta\ell_s$ be the parameter gradient, $q_s=\partial r_s/\partial p_s$ its control derivative, and $\mu_s$ Adam's first moment. In AdamW, $\beta_1$ and $\beta_2$ are the exponential retention factors for the first and second gradient moments, respectively; values closer to unity produce slower decay and hence longer optimizer memory. Omitting Hessian feedback for this illustration, a control pulse at step $s$ gives
\begin{equation}
    \delta \mu_{s+k}
    \simeq
    (1-\beta_1)\beta_1^{k-1}q_s\,\delta p_s,
    \qquad k\ge1.
\end{equation}
The first moment therefore retains an exponentially decaying record of the pulse. With the moments and optimizer clock included in $x_t$, the AdamW update is Markovian in the augmented state; a prescribed schedule may still enter $F_t$ explicitly through $t$. Removing the moments from the state leaves history-dependent parameter dynamics.

Our AdamW configurations use $\beta_1=0.9$ and $\beta_2=0.999$. The corresponding discrete e-folding times,
\begin{equation}
    \tau_\beta=-\frac{1}{\ln\beta},
\end{equation}
are approximately $9.5$ and $999.5$ steps. The eight-step horizon therefore samples a substantial fraction of the first-moment relaxation while covering only a small fraction of the second-moment timescale. The present window is consequently more sensitive to first-moment transport than to second-moment relaxation.

\subsection{Finite-action scoring}
Having identified how a perturbation is transported through the augmented state, we now use the source-time derivatives to compare finite temporal schedules. Each candidate redistributes a fixed total domain weight across several source times. An $H$-step intervention is written as
\begin{equation}
    u=(u_0,\ldots,u_{H-1}),
    \qquad
    p_{t+j}=p_0+a u_j,
\end{equation}
where $p_0$ is the baseline domain weight and $a$ is the intervention amplitude. The neutral action is $u=0$. In the primary Math--Code library, the nonneutral schedules satisfy
\begin{equation}
    \sum_{j=0}^{H-1}u_j=0,
\end{equation}
and consequently have the same aggregate domain weight over the window. They differ only in the temporal placement of the reweighting.

For a window beginning at $t$, the full source-time derivative and the optimizer-aware immediate derivative are
\begin{equation}
    \begin{aligned}
        g_{t+j}^{\mathrm{full}}
        &=C_{t+H}\Phi(t+H,t+j+1)B_{t+j},\\
        g_{t+j}^{\mathrm{my}}&=C_{t+j+1}B_{t+j}.
    \end{aligned}
    \label{eq:full-vs-myopic}
\end{equation}
The coefficient $g_{t+j}^{\mathrm{full}}$ is the single-source term of Eq.~\eqref{eq:response}, evaluated with source time $s=t+j$ and terminal time $t+H$; equivalently, $g_{t+j}^{\mathrm{full}}=\partial\J_{t+H}/\partial p_{t+j}$. It contains no summation because it describes the terminal response to a perturbation at one source time. The sum appears when the $H$ source-time contributions are contracted with a complete schedule in Eq.~\eqref{eq:schedule}. The immediate coefficient is the one-step case, for which $\Phi(t+j+1,t+j+1)=I$.
Here $C_{t+j+1}$ is the gradient of the same controller objective evaluated at the next state. The two derivatives share the source term from the current AdamW update. The immediate derivative is read out at that next state, whereas full transport carries the same source through the remaining specified updates and uses the common terminal objective. Their comparison isolates the effect of the remaining training window on the decision obtained from the current-step response.

Following Eq. \eqref{eq:hypergradient-equivalence}, the corresponding first-order terminal score can be expressed as
\begin{equation}
    \delta \J_{t+H}(u)
    =
    a\sum_{j=0}^{H-1}g_{t+j}^{\mathrm{full}}u_j.
    \label{eq:schedule}
\end{equation}
Equal-exposure schedules can therefore receive different scores when the response varies across source times.

The executed schedules have finite amplitude, and their tangent scores are used as local decision statistics. Every candidate is evaluated on the same committed future minibatches. A clipping-branch screen removes comparisons that leave the sampled smooth branch of the piecewise-smooth update~\cite{dibernardo2008piecewise}; its finite-amplitude motivation and sampled branch test are detailed in Supp.~\ref{sup:sub-remainder-screen}. In the primary seven-action experiment, the tangent-selected branch-compatible schedule is executed directly. For larger libraries, one reverse adjoint supplies the source-time derivatives, the resulting scores define a shortlist, and exact finite-amplitude rollouts select an action within that shortlist, as specified in Supp.~\ref{sup:sub-shortlist-calibration}.

\section{Experimental design}
\label{sec:experiments}

The response construction above yields a full-transport score, a memory-deleted diagnostic, and an optimizer-aware immediate comparator for exposure-matched schedules. We test these quantities prospectively in two systems, the Math--Code Transformer and the Ising--CNN model. In each case, the future minibatch path is fixed before scoring, and actions are locked before independent evaluation.

\subsection{Math--Code Transformer}

The primary experiments use a byte-level causal Transformer~\cite{vaswani2017attention} trained on paired Math and Code minibatches. Math examples come from 56 medium-entropy modules of the DeepMind Mathematics Dataset~\cite{saxton2019mathematical}; Code examples are complete files from the Python standard library~\cite{python312stdlib}. Training and validation files are disjoint, and the two domains have matched byte-token budgets; corpus construction and split definitions are given in Supp.~\ref{sup:sub-mathcode-system}. The primary model has 296,504 parameters, sequence length 64, and batch size 8, and is trained with AdamW~\cite{loshchilov2019adamw}. A 1,055,648-parameter model is used for the scale check. Normalized-age checkpoint matching and the tested duration boundary are reported in Supp.~\ref{sup:sub-mathcode-system} and Fig.~\ref{fig:s-scale-horizon}.

Control is applied in windows of $H=8$ updates. Six zero-sum schedules with amplitude $a=0.02$ compete with the neutral action, using the symmetric loss $e$ as the objective. Nonoverlapping token positions divide the validation corpus into controller, audit, and test thirds, separating action selection from the reported terminal effect. After each eight-step block, a new action is computed from the resulting checkpoint, giving blockwise receding-horizon replanning~\cite{mayne2000mpc}.

Confirmation uses 12 held-out 0.3M training histories that played no role in method development. Full transport is compared with neutral training, validation-gradient alignment (VGA), and the optimizer-aware immediate derivative in Eq.~\eqref{eq:full-vs-myopic}. The immediate derivative includes the moment update, bias correction, weight decay, and sampled clipping branch of the current AdamW step, and ends before subsequent state propagation. Checkpoints, candidate schedules, amplitudes, future minibatches, branch screens, and readout partitions are identical across arms; the complete four-arm protocol is specified in Supp.~\ref{sup:sub-mathcode-system}. Actions and terminal states are locked before token-disjoint evaluation.

\subsection{Ising--CNN benchmark}

The second system is a 252,642-parameter GroupNorm CNN~\cite{wu2018group} trained on configurations of the two-dimensional Ising model generated by Wolff-cluster Monte Carlo~\cite{wolff1989}. Relative to Math--Code, this benchmark changes the architecture, data-generating process, and scientific readout while retaining the same optimizer-state perturbation. Neural classifiers and crossing constructions have previously been used to diagnose phases and transitions~\cite{carrasquilla2017,vannieuwenburg2017confusion,bachtis2020mapping}.

The controlled neural observable, $\widehat T_N$, is the zero-crossing temperature of a linear fit to the mean logit difference. If $\bar z_{\theta}^{\mathrm{fit}}(T)=\alpha+\beta T$, then $\widehat T_N=-\alpha/\beta$, the temperature at which the fitted mean logit contrast between the two phase labels vanishes. Binder-cumulant crossings over several lattice sizes define an independent finite-size anchor $T_B$, which is locked before neural action selection~\cite{binder1981,kastening2012}. The exact infinite-lattice critical temperature obtained from Kramers--Wannier duality and the Onsager solution is reserved for post-lock evaluation~\cite{kramers1941,onsager1944}. During control, $T_B$ supplies the reference physical measurement and $\widehat T_N$ the differentiable training readout. Details of the crossing fit, Binder construction, resampling, and model averaging are given in Supp.~\ref{sup:sub-ising-system}.

And then three analyses are performed in the Ising system: an optimizer-memory ablation, a prospective control test against the locked Binder anchor, and a crossed checkpoint--future-path audit. In the crossed design, a \emph{future tape} is a pre-generated sequence of eight paired minibatches; varying the checkpoint and tape independently separates state, future-drive, and interaction contributions to schedule preference.

\subsection{Prospective protocol and statistical inference}

The two systems serve complementary roles. Math--Code supplies the primary decision-level confirmation and scale check, while Ising tests the same source--transport--readout mechanism with a different architecture, data source, and physically anchored readout.

Development and confirmation histories remain separate throughout. Predictions and branch screens are locked before finite differences or independent readouts are examined. The independently seeded training history is the primary inferential unit; windows from the same history are repeated conditions. The corresponding experimental definitions and estimands are given in Supps.~\ref{sup:sec-systems}--\ref{sup:sec-statistics}. Replication across histories is summarized by history-level bootstrap intervals and exact sign tests~\cite{efron1979}, with block resampling used to retain within-chain or contiguous-token dependence~\cite{kunsch1989}.

\section{Results}
\label{sec:results}

With the response construction and prospective protocols in place, we now examine finite-time transport in the Math--Code and Ising systems. We begin with its effect on prospectively locked decisions, then resolve the roles of the future minibatch path and optimizer-state coordinates, and finally evaluate transport-based ranking of finite-amplitude schedules.

\subsection{Decision value of delayed transport}
\label{sec:results-transport}

\begin{figure*}[t]
  \centering
  \includegraphics[width=0.98\textwidth]{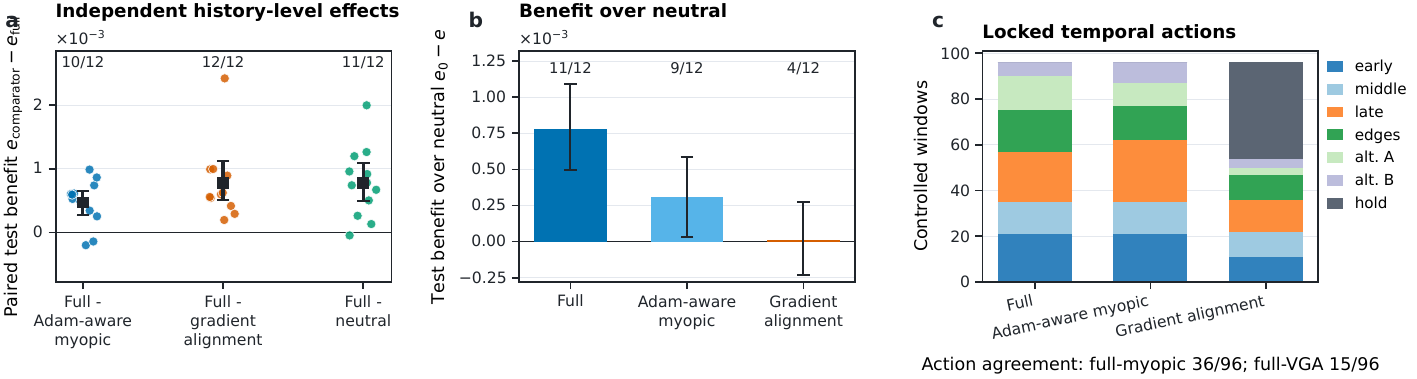}
  \caption{\textbf{Delayed transport changes finite actions and improves the independent readout.} (a) Token-disjoint history-level benefit of full eight-step transport relative to the optimizer-aware immediate derivative, VGA, and neutral training. (b) Paired benefit relative to neutral training. (c) Both full and immediate controllers intervene in all 96 windows, yet select the same temporal schedule in only 36. Actions and terminal states are locked before the test readout.}
  \label{fig:four-arm}
\end{figure*}

Using the prospective protocol above, we quantify the effect of propagation beyond the current AdamW update on locked held-out decisions. Across 12 held-out 0.3M Math--Code histories, full eight-step transport outperforms the optimizer-aware immediate controller in 10/12 histories (Fig.~\ref{fig:four-arm}). The mean token-disjoint benefit is $4.708\times10^{-4}$; its history-bootstrap 95\% interval is $[2.660,6.545]\times10^{-4}$, and the exact one-sided sign-test probability is $79/4096=0.0193$. Full transport also outperforms neutral training in 11/12 histories, with mean benefit $7.805\times10^{-4}$, and VGA in 12/12.

Both controllers intervene in all 96 windows, yet they select the same temporal schedule in only 36. Their outcome difference therefore arises from schedule selection rather than action frequency. The full-versus-immediate increment is 60.3\% of the full-versus-neutral benefit. A post-lock evaluation over essentially all available target tokens retains the sign and magnitude of the effect, with positive differences in 11/12 histories. The domain-resolved token audit and same-state decision geometry are reported in Supp.~\ref{sup:sub-four-arm} and Fig.~\ref{fig:s-decision-geometry}; the latter shows that the schedule rankings differ even when both scores are evaluated at the same state.

These comparisons isolate the decision value of propagation beyond the current AdamW update. Both controllers include the same optimizer-aware immediate derivative and intervene in every control window; they differ in how the perturbation is transported and read out over the remainder of the specified path. The resulting schedule changes and positive held-out effect show that delayed propagation contributes to prospective action quality. Because exposure-matched candidates often have similar terminal losses, this contribution can move the selected schedule across a discrete decision boundary while producing a small absolute loss difference.

The 1M extension separates local response fidelity from closed-loop decision reliability. Its pathwise tangent remains accurate, while repeated action selection gives a positive mean and 9/12 positive histories, below the prespecified 10/12 history-consistency criterion. These measurements probe different regimes: reconstruction concerns one eight-step neighborhood, whereas closed-loop utility accumulates finite-amplitude remainders and policy-induced state shifts over successive windows. A nested duration study observes this progressive separation, with model scale and suffix length varied together. The 1M experiment therefore supports the one-window response mechanism; decision-level confirmation comes from the 0.3M cohort. The reconstruction and scale evidence are documented in Supps.~\ref{sup:sub-response-scale} and \ref{sup:sec-boundaries} and Fig.~\ref{fig:s-scale-horizon}.

\subsection{Future-path conditioning}
\label{sec:results-path}

\begin{figure*}[t]
  \centering
  \includegraphics[width=0.92\textwidth]{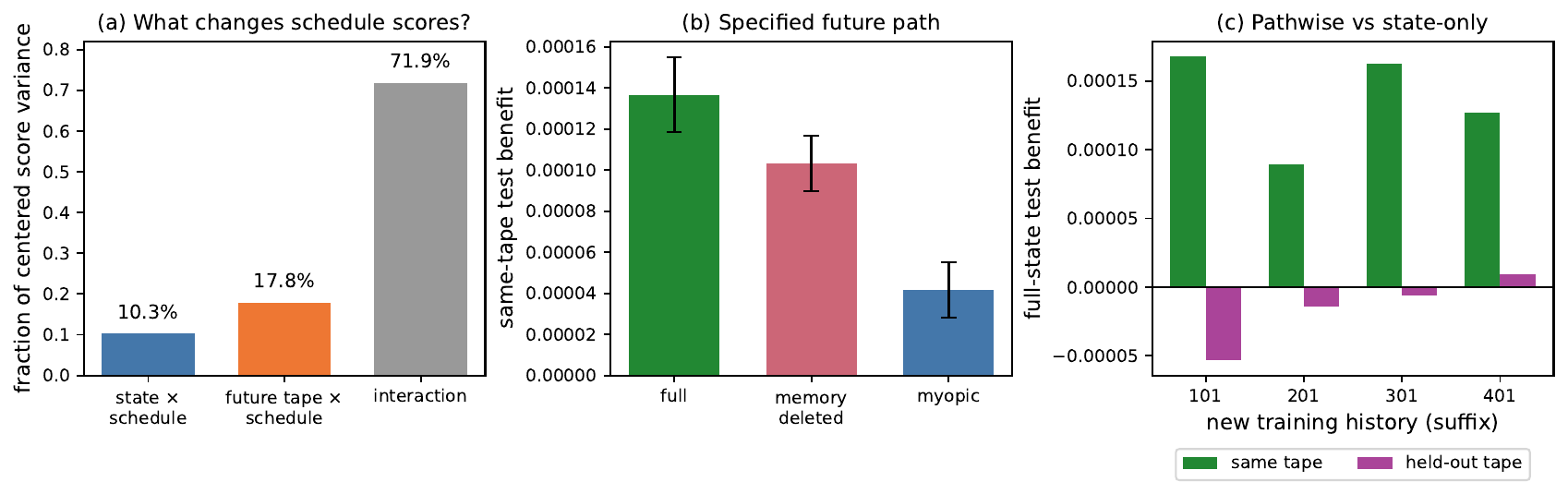}
  \caption{\textbf{The preferred short-horizon action depends on the future minibatch path.} Here a tape is a pre-generated sequence of eight paired future minibatches. (a) Schedule-score variation is dominated by checkpoint--tape interaction. (b) On the realized tape, full transport retains positive decision value relative to memory-deleted and immediate controllers. (c) The value collapses when actions are predicted from other tapes, exposing the boundary of state-only control.}
  \label{fig:factorial-audit}
\end{figure*}

Building on the Math--Code decision result, the crossed Ising design separates the contribution of the checkpoint from its interaction with future minibatches. In the resulting decomposition, centered schedule-score variance comprises 10.3\% state--schedule, 17.8\% tape--schedule, and 71.9\% state--tape--schedule interaction (Fig.~\ref{fig:factorial-audit}). The large interaction shows that the effect of a future-minibatch sequence depends strongly on its starting checkpoint.

Consistent with this interaction, a state-only leave-one-tape-out controller does not transfer reliably: its mean benefit is $-1.61\times10^{-5}$, with 1/4 positive histories. Because most schedule-score variation resides in the state--tape--schedule interaction, averaging over other tapes suppresses the path-specific component needed to rank actions on the held-out tape. By contrast, when the next eight paired minibatches are committed before action selection, full control improves 60/72 Ising cells and all six history means. The crossed and specified-path protocols are defined in Supp.~\ref{sup:sub-future-designs}, and the complete comparator results appear in Supp.~\ref{sup:sub-future-conditioning} and Fig.~\ref{fig:s-known-tape}. For the tested setting, the decision rule takes the form
\begin{equation}
    u^*=u^*(x_t,\xi_{t:t+H}),
\end{equation}
so the checkpoint alone is insufficient to determine the preferred action: the relevant control state comprises both the augmented optimizer state and the specified near-future drive.

\subsection{Optimizer-moment contributions to the response}
\label{sec:results-memory}

\begin{figure*}[t]
  \centering
  \includegraphics[width=0.92\textwidth]{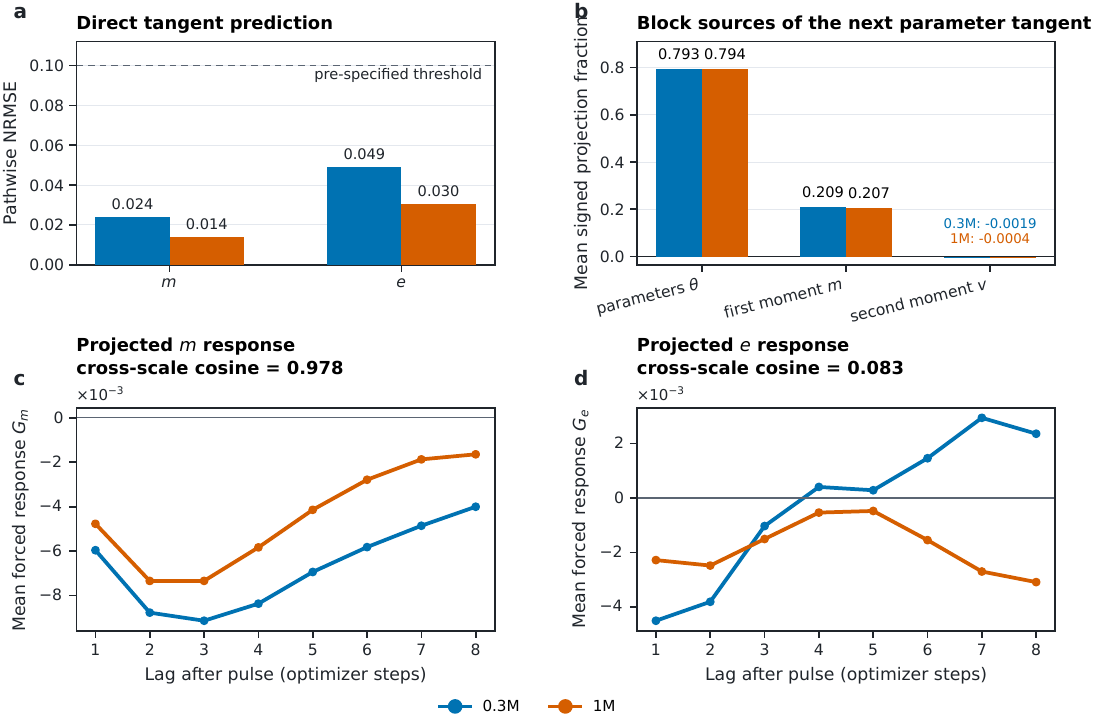}
  \caption{\textbf{Optimizer moments carry delayed response across Transformer scales.} (a) Direct pathwise tangent reconstruction remains accurate at approximately 0.3M and 1M parameters. (b) Mean signed block-projection fractions quantify contributions to the next parameter tangent; the definition and closure check are given in Supp.~\ref{sup:sub-coordinate-transport}. (c,d) Individual-path transport remains accurate even when an ensemble-mean response profile transfers weakly across scales.}
  \label{fig:mathcode}
\end{figure*}

To identify the components that carry the path-conditioned response, we resolve the perturbation across the parameter and Adam-moment coordinates of the augmented state. In Math--Code, differentiation through this full state reconstructs paired finite-amplitude responses without a fitted response model. Reconstruction is measured by the normalized root-mean-square error (NRMSE) and cosine similarity between predicted and observed response vectors. At 0.3M parameters, the full-state prediction gives $\NRMSE=0.0239$ and cosine similarity 0.9997 for $m$, and $\NRMSE=0.0490$ and cosine similarity 0.9988 for $e$. At 1M, the corresponding NRMSE/cosine pairs are 0.0138/0.9999 and 0.0304/0.9995 (Fig.~\ref{fig:mathcode}a).

Within this reconstruction, the signed block-projection fraction defined in Supp.~\ref{sup:sub-coordinate-transport} resolves contributions to the next-parameter tangent by state coordinate. The first Adam moment contributes 0.209 at 0.3M and 0.207 at 1M. In a separate optimizer intervention, reducing $\beta_1$ lowers this projection to 0.424 of baseline, whereas reducing $\beta_2$ increases the second-moment norm ratio by a factor of 20.7. The full tangent outperforms its memory-deleted counterpart in all 62 eligible momentum and AdamW-family cells. Under SGD, the same nominal pulse produces a substantially larger finite response and an inaccurate local reconstruction. The optimizer-family results and this amplitude boundary are documented in Supps.~\ref{sup:sub-response-scale} and \ref{sup:sec-boundaries}.

Ensemble averaging provides a separate test of path specificity (Fig.~\ref{fig:mathcode}c,d). Averaging paths before comparing scales gives cosine similarity 0.978 for $m$ and 0.083 for $e$, although individual 1M $e$ paths remain accurately reconstructed ($\NRMSE=0.0304$). The pooled profile combines scale dependence in the terminal projection $C$ and the transported tangent $\Phi B$, so weak transfer of the averaged $e$ profile can coexist with accurate prediction on each path. Together with the coordinate interventions, this contrast identifies optimizer moments as a measurable path-level transmission channel; the ensemble-averaged response is not a scale-independent kernel.

\subsection{Ising--CNN validation of moment-state transport}

The Math--Code analysis identifies moment-mediated transport, which we further examine in the Ising--CNN system. Full-state predictions on the Ising--CNN paths give $\NRMSE=0.002115$ for $m$ and $0.001580$ for $e$. On the same nominal trajectories, deleting moment-state tangents raises these errors to 0.6923 and 0.6871 (Fig.~\ref{fig:s-ising-mechanism}). Accurate reconstruction in this setting therefore requires transmission through the moment coordinates, consistent with the degradation observed in Math--Code.

\begin{figure*}[t]
  \centering
  \includegraphics[width=0.92\textwidth]{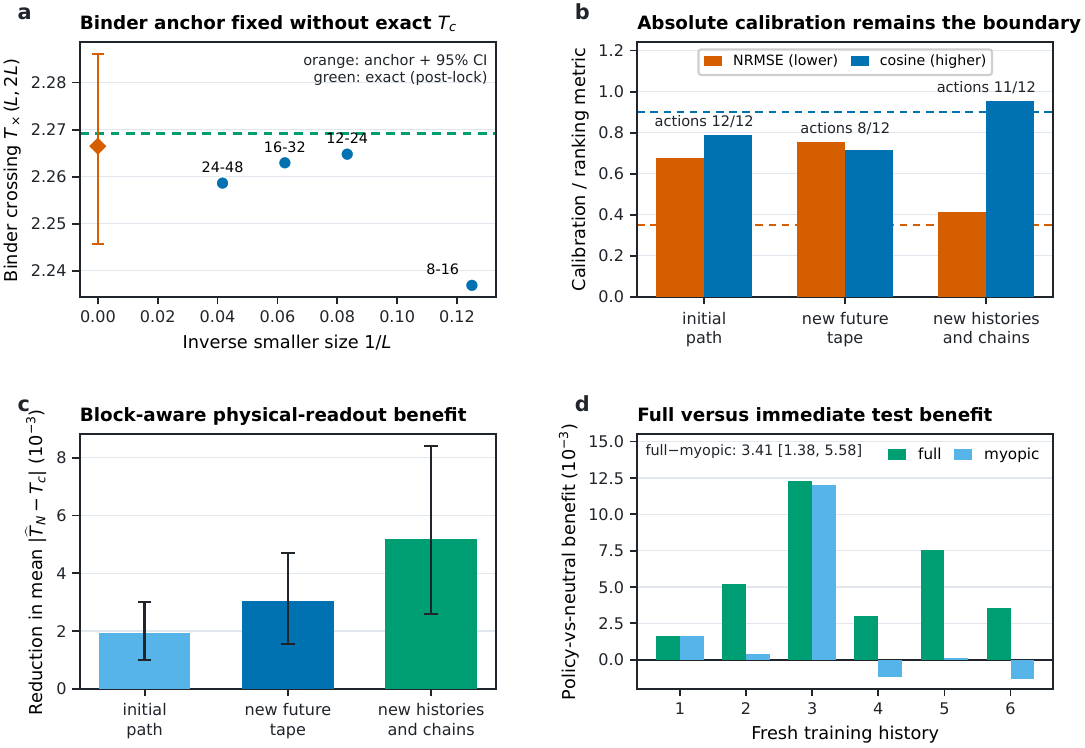}
  \caption{\textbf{The optimizer-state mechanism transfers to a Binder-anchored Ising benchmark.} (a) Binder crossings define the physical anchor before neural action selection; the final confirmation reuses this anchor while replacing neural paths and test chains. (b) Candidate-contrast NRMSE, cosine similarity, and discrete action agreement across the staged tests. The elevated absolute-calibration NRMSE marks a calibration boundary, while action agreement remains high. (c) The selected action reduces the absolute deviation of the neural crossing from the Binder anchor on new test chains. (d) Full transport outperforms the immediate controller relative to neutral training.}
  \label{fig:t8-anchor}
\end{figure*}

Beyond response reconstruction, the prospective Ising test uses the independently locked Binder anchor. Its final confirmation retains the fixed $T_B$ and replaces the neural histories, future tapes, calibration chains, and test chains. Full and exhaustive-rollout policies agree in 11/12 cells. On the new test chains, the mean absolute deviation $|\widehat T_N-T_B|$ falls from 0.02435 to 0.01880. All six histories improve, with a block-aware interval of $[0.00258,0.00839]$ (Fig.~\ref{fig:t8-anchor}). Binder finite-size analysis remains the critical-point reference, while the neural crossing serves as the controlled training readout. The calibration boundary and fresh-chain confirmation are reported in Supp.~\ref{sup:sub-binder-control}; the anchored protocol is defined in Supp.~\ref{sup:sub-ising-system}. Together, these results reproduce moment-mediated transport under a different architecture and readout, with the Binder construction providing an external physical reference for the controlled neural observable.

\subsection{Transport-guided ranking of finite actions}
\label{sec:results-ranking}

With the local mechanism established, we evaluate transport scores as a way to reduce exact finite-amplitude search in larger schedule libraries.
For a scalar terminal objective, one reverse adjoint gives derivatives at every source time. A schedule library can then be ranked by scalar products. Because finite amplitude may change the ordering, exact rollouts evaluate the shortlisted candidates. Ranking quality is measured by \emph{winner recall}: the fraction of settings in which the exhaustive finite-amplitude winner lies among the top-$q$ candidates of a ranking rule.

\begin{figure*}[t]
  \centering
  \includegraphics[width=0.98\textwidth]{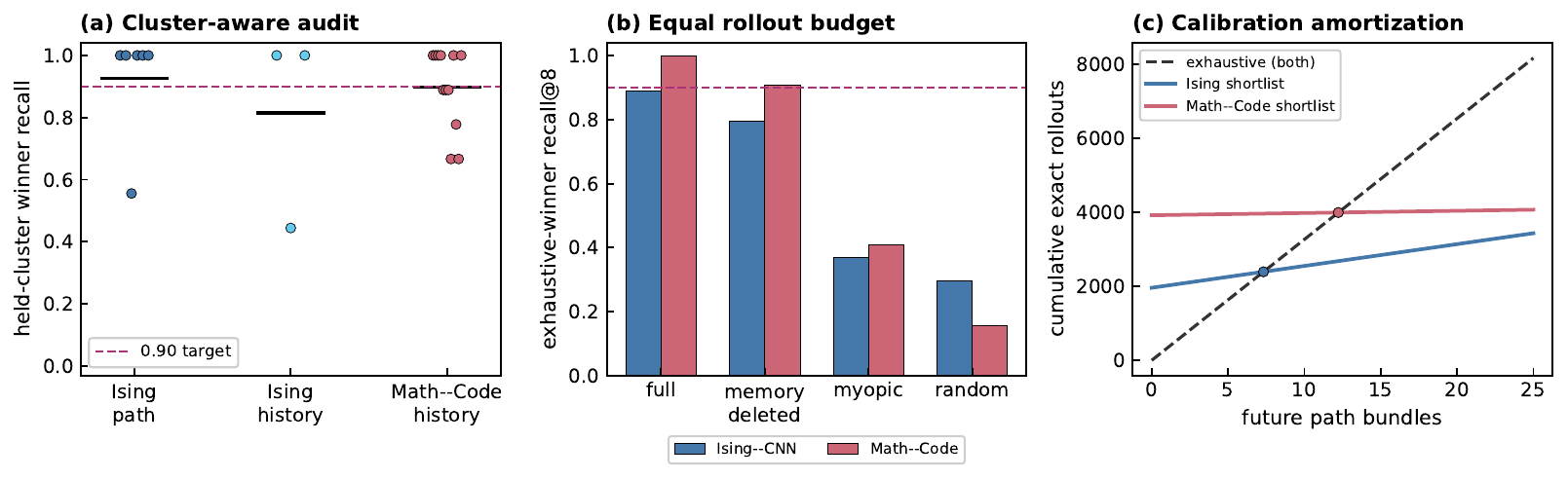}
  \caption{\textbf{Full transport concentrates finite-amplitude winners at equal exact-rollout budget.} (a) Each colored point is the winner recall obtained after holding out one Ising path, Ising history, or Math--Code history cluster; black segments denote the mean within each cluster type. (b) With neutral plus $q=8$ nonneutral rollouts, full transport has the highest exhaustive-winner recall in both systems, followed by memory deletion, immediate ranking, and random selection. (c) Charging the one-time development calibration gives break-even points of 7.32 Ising and 12.22 Math--Code future-path bundles for repeated reuse of the same library structures. This accounting counts candidate rollouts rather than total training time.}
  \label{fig:cost}
\end{figure*}

On 54 fresh Ising settings, a development-calibrated adaptive shortlist includes the exhaustive finite-amplitude winner in every case, and exact reranking selects it without a clipping-branch mismatch (Fig.~\ref{fig:cost}). At an equal budget of neutral plus eight nonneutral exact rollouts, winner recall for full, memory-deleted, immediate, and random ranking is 0.889, 0.796, 0.370, and 0.296. The corresponding Math--Code values are 1.000, 0.907, 0.407, and 0.155. With fixed $q=16$, exact reranking recovers all 108 exhaustive winners on fresh Math--Code settings. The adaptive rule recovers 100/108 overall; one library gives 32/36, compared with its prespecified criterion of 33/36. The missed candidates occur where the finite-amplitude remainder and local score spacing exceed the calibrated cutoff, quantities not determined by amplitude and library size alone. Its locked actions still improve the token-disjoint objective in all six fresh histories, with mean benefit $6.46\times10^{-4}$. The transfer audit and calibration boundary are reported in Supp.~\ref{sup:sub-ranking-cost} and Fig.~\ref{fig:s-mathcode-transfer}; measured costs appear in Tables~\ref{tab:s-four-arm-cost}--\ref{tab:s-complete-cost}.

These results support full transport as a screening statistic that concentrates strong finite actions, while exact rollout reranking resolves finite-amplitude competition within the shortlist. The fixed-$q$ result is stable across the tested libraries, whereas the adaptive width remains system-calibrated.

\section{Discussion}
\label{sec:discussion}

Across the two systems, short-horizon loss-weight control emerges as a path-conditioned dynamical problem on the augmented model--optimizer state. Delayed propagation changes prospectively locked decisions, Adam moments carry a measurable part of the perturbation, and the preferred schedule depends jointly on the current optimizer state and the near-future drive. These findings also distinguish accurate local response prediction from reliable finite-amplitude action selection.

These results connect recent analyses of noncommuting domain updates, optimizer-memory order effects, and optimizer-aware attribution to prospective decisions over finite temporal actions~\cite{rukhovich2025commute,sweeney2026optmemory,ding2026adamshapley}. They also complement truncated hypergradients~\cite{shaban2019truncated} and dynamic data-mixture methods that modify sample allocation or the subsequent training path~\cite{xie2023doremi,fan2024doge,fan2024dga,li2025pike}. In the paired-loss setting studied here, the differentiated object is a specified finite future map. Aggregate domain exposure remains fixed, so the action changes the temporal placement of domain weights rather than their total allocation. The Ising benchmark tests the same dynamical mechanism in a different architecture, data-generating system, and physically anchored readout.

Finite amplitude introduces an additional ranking problem because curvature can reorder candidates that are close under the local tangent. The small candidate library therefore tests the tangent-selected action directly, whereas the larger libraries use full transport to form a shortlist before exact rollout reranking. At equal exact-rollout budget, full transport places the exhaustive winner near the top more often than immediate, memory-deleted, or random ranking. The adaptive shortlist requires system-specific empirical calibration and therefore complements, rather than replaces, fixed-budget shortlist selection. The computational tradeoff depends on how the derivatives are evaluated. Forward full transport propagates source sensitivities through the augmented state, whereas the reverse adjoint obtains all source-time derivatives in one backward sweep while retaining intermediate trajectory states. In the measured CPU implementation, forward full transport takes 1.33 times the optimizer-aware immediate-controller total; reverse-adjoint full transport takes 0.80 times that total but 2.72 times its peak resident memory. These measurements explain the observed time--memory tradeoff and identify the regime in which transport-guided screening is most useful: candidate libraries large enough to benefit from shortlist reranking, especially when their structure is reused across future paths.

The present evidence defines a local operating regime for this description. It is strongest for eight-step control with Adam-like optimizers and the 0.3M Transformer. At 1M parameters, the mean decision effect remains positive, while accumulated finite-amplitude and closed-loop deviations accompany a missed history-consistency criterion. Under SGD, the common pulse produces a much larger response, confounding optimizer family with effective perturbation scale. The assumption of a specified near-future minibatch sequence is also substantive. Unknown future drives would require averaging over possible paths, while changes in sample allocation would introduce additional gradient-covariance effects. Larger models, longer horizons, amplitude-matched optimizer comparisons, and future-path marginalization provide direct tests of how far the finite-time description extends.

\section{Conclusion}

Adaptive optimizers retain gradient history, but how this internal state changes finite training decisions has been unclear. We addressed this problem by differentiating specified eight-step AdamW paths through the complete model--optimizer state, selecting exposure-matched temporal schedules, and locking each action before independent evaluation. Math--Code supplied the primary prospective decision test, while an Ising--CNN benchmark examined the same mechanism under a different architecture, data-generating system, and physically anchored readout.

In the primary 0.3M Math--Code confirmation, full transport improved held-out decisions relative to an optimizer-aware immediate derivative and reordered schedules despite matched intervention frequency. Crossed checkpoint--path experiments showed that the preferred action depends jointly on the current optimizer state and the near-future minibatches, while targeted tangent deletion identified Adam moments as a measurable transmission channel in both systems. For larger candidate libraries, transport scores concentrated exact-rollout winners and focused finite-amplitude evaluation on a shortlist.

These results identify short-horizon loss-weight control as a path-conditioned dynamical problem on the augmented training state. Along committed paths, full-state derivatives can therefore serve both as mechanistic probes of training dynamics and as screening scores for finite interventions. Future-path marginalization, longer horizons, and larger models provide direct tests of how far this local description extends.

\begin{acknowledgments}
The work of J.G. is supported by the Postdoctoral Fellowship Program (Grade C) of China Postdoctoral Science Foundation under Grant No. GZC20252775. The author used OpenAI ChatGPT and Codex to assist with code debugging, literature searches, and language editing. All AI-generated suggestions were reviewed by the author, who assumes full responsibility for the content of this work.
\end{acknowledgments}

\section*{Data Availability and Code}

The code required to reproduce the analyses in this work is publicly available in the \href{https://github.com/gitguojh/Delayed-Optimizer-State-Transport-Shapes-Short-Horizon-Training-Decisions}{project's GitHub repository}. All relevant input parameters and plotting scripts are included there.

\clearpage
\onecolumngrid
\raggedbottom
\begin{center}
  {\large\bfseries Supplemental Material for \\ Delayed Optimizer-State Transport Shapes Short-Horizon Training Decisions \par}
  \vspace{0.8em}
  {\normalsize Jinhui Guo\par}
  \vspace{0.25em}
  {\small School of Physics, Beihang University, Beijing 100191, China\par}
\end{center}
\vspace{0.8em}

\setcounter{section}{0}
\setcounter{subsection}{0}
\renewcommand{\thesection}{\Alph{section}}
\renewcommand{\thesubsection}{\thesection.\arabic{subsection}}
\makeatletter
\renewcommand{\p@subsection}{}
\makeatother
\setcounter{equation}{0}
\renewcommand{\theequation}{S\arabic{equation}}
\renewcommand{\theHsection}{supplement.\arabic{section}}
\renewcommand{\theHequation}{supplement.\arabic{equation}}
\setcounter{figure}{0}
\renewcommand{\thefigure}{S\arabic{figure}}
\renewcommand{\theHfigure}{supplement.\arabic{figure}}
\setcounter{table}{0}
\renewcommand{\thetable}{S\arabic{table}}
\renewcommand{\theHtable}{supplement.\arabic{table}}

This Supplemental Material provides detailed derivations, experimental definitions, statistical procedures, additional results, and reproducibility information supporting the main text. The notation follows that of the main text. Supp.~\ref{sup:sec-response-theory} gives the AdamW-specific tangent equations and coordinate-resolved comparator implementation. Supps.~\ref{sup:sec-systems}--\ref{sup:sec-finite-actions} specify the experimental systems, statistical estimands, branch safeguards, and finite-action protocol. Supp.~\ref{sup:sec-supporting-results} presents the supporting results and cost measurements, while Supps.~\ref{sup:sec-boundaries} and \ref{sup:sec-reproducibility} summarize the empirical boundaries and reproducibility record.

Throughout, \emph{locked before evaluation} means that the configuration, implementation, candidate set, predictions or actions, and any terminal-state hashes were fixed before the designated audit or independent test outcome was examined. This ordering keeps the stated comparisons outcome-blind. The lock records are internal protocol records and were not externally preregistered.

\FloatBarrier
\suppsection{AdamW optimizer-state tangent and coordinate transport}{sup:sec-response-theory}

The source--transport--readout factorization and finite-action scores are defined in Sec.~\ref{sec:response-theory}. This section supplies the AdamW-specific tangent equations, coordinate decomposition, and implementation checks used for optimizer-memory attribution.

\subsection{AdamW tangent equations}
\label{sup:sub-adamw-tangent}

For AdamW~\cite{kingma2014adam,loshchilov2019adamw}, let $r_t=\nabla_\theta\ell_t(\theta_t,p_t)$ be the raw gradient and let $\widetilde r_t=\mathcal C_t(r_t)$ denote the gradient after the active clipping map. The variables $\mu_t$ and $v_t$ are the uncorrected exponential moving averages of the clipped gradient and its elementwise square, respectively. Here $\odot$ and $\oslash$ denote elementwise multiplication and division, and square roots of vector-valued moment quantities are also evaluated elementwise. The moment updates are
\begin{equation}
\begin{aligned}
 \mu_{t+1}&=\beta_1\mu_t+(1-\beta_1)\widetilde r_t,\\
 v_{t+1}&=\beta_2v_t+(1-\beta_2)\widetilde r_t\odot\widetilde r_t.
\end{aligned}
 \label{eq:s-adam-moments}
\end{equation}
If $n_{t+1}$ is the optimizer clock after the update, the hats denote bias-corrected moments,
\begin{equation}
 \widehat\mu_{t+1}=\frac{\mu_{t+1}}{1-\beta_1^{n_{t+1}}},
 \qquad
 \widehat v_{t+1}=\frac{v_{t+1}}{1-\beta_2^{n_{t+1}}}.
 \label{eq:s-bias-correction}
\end{equation}
With $d_{t+1}=\sqrt{\widehat v_{t+1}}+\epsilon$, the parameter update is
\begin{equation}
 \theta_{t+1}=(1-\eta_t\lambda)\theta_t
 -\eta_t\widehat\mu_{t+1}\oslash d_{t+1}.
 \label{eq:s-adam-parameter}
\end{equation}
Here $\eta_t$ is the learning rate, $\lambda$ is the decoupled weight-decay coefficient, and $\epsilon$ is the denominator stabilizer.

On a fixed differentiable clipping branch, let $H_t=\partial r_t/\partial\theta$ and $q_t=\partial r_t/\partial p$. Then
\begin{equation}
 \delta\widetilde r_t
 =D\mathcal C_t(r_t)\left(H_t\delta\theta_t+q_t\delta p_t\right).
 \label{eq:s-clipped-gradient-tangent}
\end{equation}
For the paired weighted loss, $q_t=\nabla_\theta\ell_{A,t}-\nabla_\theta\ell_{B,t}$. The moment tangents are
\begin{equation}
\begin{aligned}
 \delta\mu_{t+1}&=\beta_1\delta\mu_t+(1-\beta_1)\delta\widetilde r_t,\\
 \delta v_{t+1}&=\beta_2\delta v_t
 +2(1-\beta_2)\widetilde r_t\odot\delta\widetilde r_t.
 \label{eq:s-adam-tangent}
\end{aligned}
\end{equation}
Because the control does not perturb the optimizer clock,
\begin{equation}
 \delta\widehat\mu_{t+1}=\frac{\delta\mu_{t+1}}{1-\beta_1^{n_{t+1}}},
 \qquad
 \delta\widehat v_{t+1}=\frac{\delta v_{t+1}}{1-\beta_2^{n_{t+1}}}.
\end{equation}
The parameter tangent is therefore
\begin{equation}
\begin{aligned}
 \delta\theta_{t+1}=(1-\eta_t\lambda)\delta\theta_t -\eta_t\left[
 \delta\widehat\mu_{t+1}\oslash d_{t+1}
 -\left(\widehat\mu_{t+1}\odot\delta\widehat v_{t+1}\right)
 \oslash\left(2\sqrt{\widehat v_{t+1}}\odot d_{t+1}\odot d_{t+1}\right)
 \right].
 \label{eq:s-parameter-tangent}
\end{aligned}
\end{equation}
The implementation evaluates the clipping differential and Hessian--vector products by matrix-free automatic differentiation~\cite{pearlmutter1994hvp}; no dense Hessian is formed.

\subsection{Coordinate-resolved transport and comparator implementation}
\label{sup:sub-coordinate-transport}

The equations above propagate the complete state tangent. To resolve the contribution of individual state coordinates, let $P_j$ project onto $j\in\{\theta,\mu,v\}$, let $\dot x_k$ and $\dot p_k$ denote the state and control tangents along the specified intervention direction, and write $\dot x_{i,k+1}=P_i\dot x_{k+1}$. At a propagation step after the source injection, where $\dot p_k=0$, the contribution from input block $j$ to output block $i$ after one update is

\begin{equation}
 w_{i\leftarrow j,k}=P_iA_kP_j\dot x_k,
 \qquad
 c_{i\leftarrow j,k}=\frac{\langle\dot x_{i,k+1},w_{i\leftarrow j,k}\rangle}{\|\dot x_{i,k+1}\|^2}.
 \label{eq:s-block-projection}
\end{equation}
If a control is applied at step $k$, $P_iB_k\dot p_k$ enters separately as a new source contribution.
The denominator in Eq.~\eqref{eq:s-block-projection} is the squared norm of the target tangent, so $c_{i\leftarrow j,k}$ is a signed reconstruction contribution. For $i=\theta$, the mean $(\theta,\mu,v)$ contributions are $(0.7930,0.2089,-0.00187)$ at 0.3M and $(0.7936,0.2068,-0.00037)$ at 1M. Their near-unit sums give an internal closure check. In the 1M implementation audit, the maximum block-closure and readout-factorization errors are $5.96\times10^{-8}$ and $1.54\times10^{-8}$, respectively. The approximately 21\% value in the main text is the first-moment block's signed projection onto the next parameter tangent. Supp.~\ref{sup:sub-response-scale} reports direct finite-response reconstruction; the timed forward and adjoint implementations select the same action on all four checked paths (Table~\ref{tab:s-four-arm-cost}).

The memory-deleted diagnostic projects the propagated tangent after every update as $(\delta\theta,\delta\mu,\delta v)\mapsto(\delta\theta,0,0)$ while leaving the nominal parameters, moments, and future minibatches unchanged. The optimizer-aware immediate comparator uses the same complete one-step tangent but terminates before subsequent propagation, as defined in Eq.~\eqref{eq:full-vs-myopic}. These two constructions separately test moment-coordinate transmission and the decision value of the remaining specified future map.

\suppsection{Experimental systems and prospective controls}{sup:sec-systems}

Section~\ref{sec:experiments} introduces the two experimental systems and their prospective tests. Here we give the model configurations, data partitions, candidate schedules, physical readouts, and crossed future-path construction needed to interpret and reproduce those tests.

\subsection{Math--Code Transformer}
\label{sup:sub-mathcode-system}

Domain $A$ contains 56 medium-entropy modules of the DeepMind Mathematics Dataset; domain $B$ contains complete Python standard-library files. Training and validation files are disjoint, with matched byte-token budgets. The causal byte-level Transformers~\cite{vaswani2017attention} use sequence length 64, batch size 8, zero dropout, four attention heads, and AdamW with $\beta_1=0.9$ and $\beta_2=0.999$. The smaller model has 296,504 parameters in three width-104 layers, and the larger model has 1,055,648 parameters in four width-176 layers.

Training progress is indexed by an effective weighted token exposure per model parameter. With $t$ optimizer steps, per-domain minibatch size $B$, sequence length $S$, and parameter count $P$, this gives
\begin{equation}
 \tau=\frac{tBS}{P},\qquad \Delta\tau_{\mathrm{control}}=\frac{H_{\mathrm{total}}BS}{P},
 \label{eq:s-normalized-age}
\end{equation}
where $tBS$ is the exposure of the unit-normalized mixture loss. Each paired step evaluates $B$ sequences from each domain, so the computational token count is $2tBS$; $\tau$ uses the weighted-mixture convention to provide a common checkpoint coordinate. Dividing by $P$ compares how much weighted data exposure each parameter has received. Steps 1750 at 0.3M and 6200 at 1M both lie near $\tau\simeq3$. The same coordinate maps a 64-step controlled suffix at 0.3M to approximately 228 steps at 1M; the long 1M development trajectory uses 224 steps.

\begin{figure}[H]
 \centering
 \includegraphics[width=0.94\textwidth]{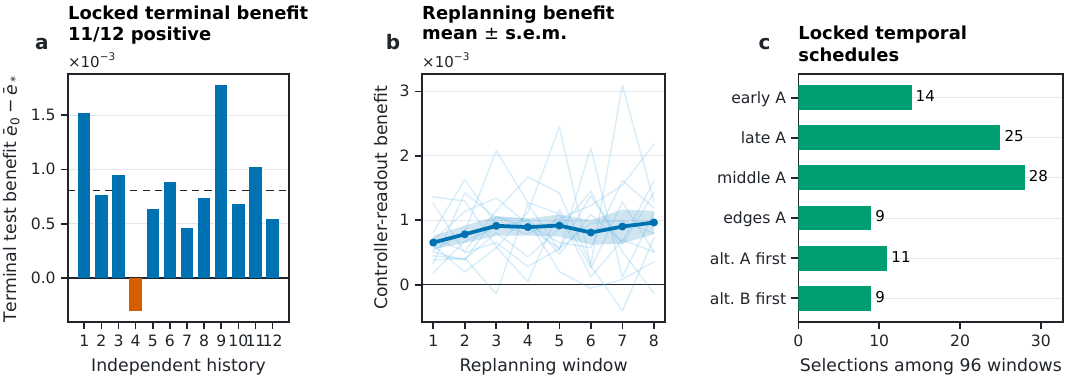}
 \caption{Receding-horizon behavior in the independent 0.3M full-versus-neutral cohort. (a) Terminal test benefit $\bar e_0-\bar e_*$ by training history; the dashed line marks the history mean. (b) Controller-readout benefit over the eight replanning windows; thin lines show histories, and the dark curve and band show the mean and its standard error. (c) Selection counts for the schedules in Eq.~\eqref{eq:s-mathcode-schedules}.}
 \label{fig:s-mathcode-control}
\end{figure}

The primary library contains neutral and six zero-sum eight-step schedules, with $u_j\in\{-1,+1\}$, $\sum_ju_j=0$, and amplitude $a=0.02$. A plus sign increases the Math weight to $p_0+a$ and a minus sign increases the Code weight to $1-p_0+a$. The schedule names used in Fig.~\ref{fig:s-mathcode-control}c correspond to
\begin{equation}
\begin{aligned}
 u^{\mathrm{early\ A}}&=(+,+,+,+,-,-,-,-),&
 u^{\mathrm{late\ A}}&=(-,-,-,-,+,+,+,+),\\
 u^{\mathrm{middle\ A}}&=(-,-,+,+,+,+,-,-),&
 u^{\mathrm{edges\ A}}&=(+,+,-,-,-,-,+,+),\\
 u^{\mathrm{alt.\ A}}&=(+,-,+,-,+,-,+,-),&
 u^{\mathrm{alt.\ B}}&=(-,+,-,+,-,+,-,+).
\end{aligned}
\label{eq:s-mathcode-schedules}
\end{equation}
All six therefore have the same aggregate Math and Code weight over a window and differ only in temporal order. Guarded token positions divide the validation corpus into controller, audit, and test thirds. The controller partition is used to score actions; audit and test readouts are evaluated after the actions and terminal-state hashes have been locked.

The primary four-arm experiment compares neutral, VGA, the Adam-aware immediate derivative, and full transport on 12 held-out histories under identical checkpoints, future minibatches, schedules, branch screens, and readout partitions; Supp.~\ref{sup:sub-four-arm} gives the full comparison. An earlier independent cohort used the same 0.3M model and schedule library to compare full transport with neutral training over eight replanning windows. Denote their terminal symmetric test losses by $\bar e_*$ and $\bar e_0$. The benefit $\bar e_0-\bar e_*$ is positive in 11/12 histories, and its mean remains positive throughout the suffix (Fig.~\ref{fig:s-mathcode-control}a,b). The selected schedules span the complete nonneutral library (Fig.~\ref{fig:s-mathcode-control}c), so the ordering is recomputed as the controlled trajectory evolves.

\begin{figure}[htbp]
 \centering
 \includegraphics[width=0.94\textwidth]{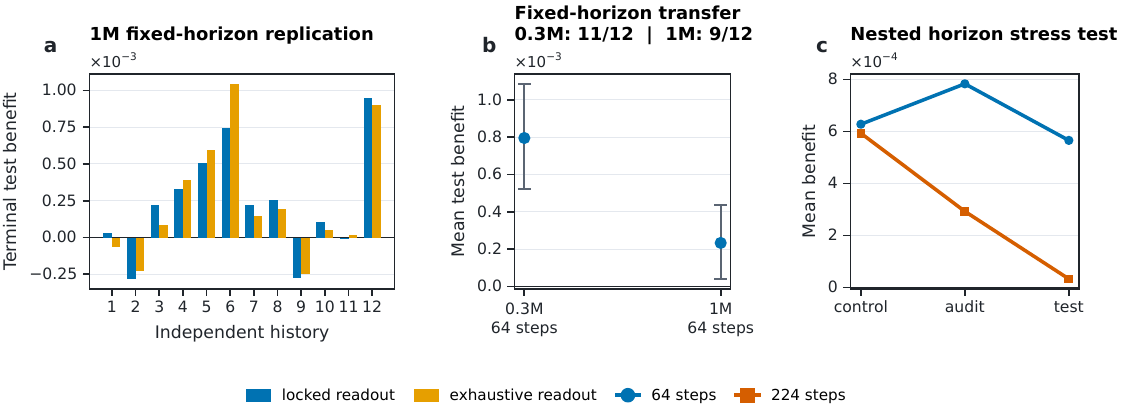}
 \caption{Math--Code scale and duration observations. (a) History-level 1M benefits on the locked and dense test readouts. (b) Mean 64-step benefit at 0.3M and 1M; error bars are history-bootstrap 95\% intervals. (c) Controller, audit, and test benefits after 64 and 224 controlled steps in one nested 1M development trajectory.}
 \label{fig:s-scale-horizon}
\end{figure}

The scale extension keeps the training age near $\tau=3$ and first retains the same 64-step suffix. Its mean benefit is $2.329\times10^{-4}$ with history-bootstrap interval $[0.399,4.384]\times10^{-4}$; 9/12 histories are positive, compared with the prespecified consistency threshold of 10/12. The dense token readout preserves this count (Fig.~\ref{fig:s-scale-horizon}a,b). In the single 224-step development trajectory, controller-readout benefit remains positive while independent-test benefit approaches zero (Fig.~\ref{fig:s-scale-horizon}c). The longer suffix accumulates finite-amplitude displacement and changes the states at which later windows are linearized. At 1M, accurate one-window response therefore coexists with less uniform utility after repeated closed-loop interventions.

\FloatBarrier
\subsection{Ising--CNN system and physical anchor}
\label{sup:sub-ising-system}

Two-dimensional Ising configurations are generated by Wolff cluster updates~\cite{wolff1989}. The supervised class label and the controlled data domain are distinct. Class 0 contains ordered-phase anchors at temperatures $\{1.50,1.80,2.10,2.18\}$, and class 1 contains disordered-phase anchors at $\{2.36,2.44,2.80,3.20\}$. Domain $A$ contains the four anchors nearer the transition, $\{2.10,2.18,2.36,2.44\}$, whereas domain $B$ contains the four farther anchors, $\{1.50,1.80,2.80,3.20\}$. The temperatures 2.24 and 2.30 are excluded from supervised training and appear only in the neural transition readout. The 252,642-parameter $L=16$ CNN uses GroupNorm~\cite{wu2018group}, zero dropout, and AdamW. Its temperature-resolved logit crossing is the differentiable training observable, following earlier machine-learning studies of phases and transitions~\cite{carrasquilla2017,vannieuwenburg2017confusion,bachtis2020mapping}.

For a configuration $\sigma$ at temperature $T$, let $z_0(\sigma;\theta)$ and $z_1(\sigma;\theta)$ be the ordered- and disordered-class logits. For $N_T$ readout configurations $\{\sigma_i^{(T)}\}_{i=1}^{N_T}$, define the mean disordered-minus-ordered contrast
\begin{equation}
 \bar z_\theta(T)=\frac{1}{N_T}\sum_{i=1}^{N_T}\left[z_1(\sigma_i^{(T)};\theta)-z_0(\sigma_i^{(T)};\theta)\right].
 \label{eq:s-logit}
\end{equation}
A line $\bar z_\theta(T)=\alpha+\beta T$ fitted at $T\in\{2.18,2.24,2.30,2.36\}$ defines $\widehat T_N=-\alpha/\beta$, the temperature at which the fitted classifier preference changes sign. This $L=16$ neural crossover is the differentiable controller readout. The Binder construction below provides the thermodynamic reference.

Let $m_{\mathrm{spin}}=L^{-2}\sum_i\sigma_i$ be the magnetization per spin. The Binder cumulant and each ratio-two crossing are defined by
\begin{equation}
 U_L(T)=1-\frac{\langle m_{\mathrm{spin}}^4\rangle_T}{3\langle m_{\mathrm{spin}}^2\rangle_T^2},
 \qquad U_L(T_\times)=U_{2L}(T_\times),
 \label{eq:s-binder}
\end{equation}
and their finite-size drift is modeled as
\begin{equation}
 T_\times(L,2L)=T_B+AL^{-r}.
 \label{eq:s-binder-drift}
\end{equation}
Here $T_B$ is the extrapolated crossing temperature, $A$ the drift amplitude, and $r$ a correction exponent. The Monte Carlo bank contains eight independent chains, 256 configurations per temperature, eight temperatures from 2.10 to 2.45, and sizes $L\in\{8,12,16,24,32,48\}$. Crossings from $(8,16)$, $(12,24)$, $(16,32)$, and $(24,48)$ are fitted at frozen $r\in\{0.5,1.0,\ldots,4.0\}$ and combined using Akaike weights~\cite{akaike1974}. Circular-block resampling propagates within-chain dependence, and variation across the fitted exponents is included as between-model uncertainty~\cite{kunsch1989}. The resulting $T_B$ distribution is frozen before neural action selection. Post-lock evaluation uses the exact infinite-lattice value $2/\log(1+\sqrt2)$~\cite{kramers1941,onsager1944}.

For eight independent $L=16$ neural calibration chains, the controlled objective is
\begin{equation}
 \J_{\mathrm{anchor}}=(\overline{\widehat T_N}-T_B)^2+\operatorname{Var}_c(\widehat T_{N,c})+0.1\frac{\overline{R_c}}{(\overline{\beta_c})^2+10^{-12}},
 \label{eq:s-anchor-objective}
\end{equation}
where $c$ indexes the eight neural calibration chains, $\widehat T_{N,c}$ is the fitted crossing on chain $c$, $\beta_c$ is its fitted slope, and $R_c$ is its mean squared residual. Overbars denote the mean across chains, and $\operatorname{Var}_c$ is the population variance across the eight chains (divisor eight). The three terms penalize displacement from the Binder anchor, variation across neural chains, and poor conditioning of the fitted crossing. The coefficient 0.1 fixes the relative weight of the fit-quality term, and $10^{-12}$ stabilizes its denominator. At baseline $p_0=0.5$, the nine-candidate library contains neutral, the six zero-sum schedules in Eq.~\eqref{eq:s-mathcode-schedules} with $A$ denoting the near-transition domain, and two constant schedules: near-heavy $=(+,+,+,+,+,+,+,+)$ and far-heavy $=(-,-,-,-,-,-,-,-)$. The anchored protocol uses $a=0.08$. Binder scaling fixes the physical reference used in $\J_{\mathrm{anchor}}$, which is differentiated to score the candidate actions.

\subsection{Crossed future-tape and specified-path designs}
\label{sup:sub-future-designs}

The two designs resolve how the current checkpoint and the next minibatches jointly determine a schedule. The crossed audit measures transfer of a checkpoint-conditioned schedule across possible future minibatch sequences. It uses four independent histories, checkpoints 60, 100, and 140, and the same eight future tapes at each of the resulting 12 states, giving $12\times8=96$ state--tape cells at amplitude $a=0.002$. A future tape is a pre-generated sequence of eight paired near- and far-domain minibatches.

Let $S_{i\nu k}$ be the full-transport score of schedule $k$ from state $i$ on tape $\nu$, and let $Z_{i\nu k}=S_{i\nu k}-S_{i\nu 0}$ be its contrast with the neutral schedule. The balanced crossed design decomposes this contrast as
\begin{equation}
 Z_{i\nu k}=\overline Z_{\cdot\cdot k}
 +\Delta^{\mathrm{state}}_{ik}
 +\Delta^{\mathrm{tape}}_{\nu k}
 +\Delta^{\mathrm{int}}_{i\nu k},
 \label{eq:s-state-tape-decomposition}
\end{equation}
where $\Delta^{\mathrm{state}}_{ik}=\overline Z_{i\cdot k}-\overline Z_{\cdot\cdot k}$ is the state--schedule term, $\Delta^{\mathrm{tape}}_{\nu k}=\overline Z_{\cdot\nu k}-\overline Z_{\cdot\cdot k}$ the tape--schedule term, and $\Delta^{\mathrm{int}}_{i\nu k}$ the remaining state--tape--schedule interaction. Each component is broadcast to the full $(i,\nu,k)$ array before its entries are squared and summed to give $SS_r$. The reported percentage for $r\in\{\mathrm{state},\mathrm{tape},\mathrm{int}\}$ is $SS_r/(SS_{\mathrm{state}}+SS_{\mathrm{tape}}+SS_{\mathrm{int}})$.

For the leave-one-tape-out test, each tape in turn is excluded. Scores and branch eligibility are aggregated over the other seven tapes, producing a checkpoint-conditioned action for the held-out tape. Comparators are neutral, memory-deleted, Adam-aware immediate, random matched-weight schedules, and a globally fixed schedule selected from the same seven tapes.

The specified-path design conditions the decision on the committed eight-minibatch tape. The same tape therefore determines the propagator and the executed finite action. Six new histories, three checkpoints per history, and four tapes per checkpoint yield 72 cells. Predictions, actions, branch records, checkpoints, and code are locked before generation of a new independent test chain. Its decision rule is $u^*(x_t,\xi_{t:t+H})$. Results for the crossed and specified-path designs are reported in Sec.~\ref{sec:results-path} and Supp.~\ref{sup:sub-future-conditioning}.

\FloatBarrier
\suppsection{Estimands and statistical methodology}{sup:sec-statistics}

Section~\ref{sec:results} reports response reconstruction, history-level control benefits, and recovery of finite-amplitude winners. This section defines the corresponding estimands and their units of replication.

Let $y_t^{(s,+a)}$ and $y_t^{(s,-a)}$ denote the same scalar or vector observable at time $t$ after changing the loss weight only at source time $s$ from its nominal value $p_0$ to $p_0+a$ or $p_0-a$. The two trajectories use identical minibatches and differ only in the sign of this perturbation. Their central finite-amplitude response is
\begin{equation}
 r_y(t,s;a)=\frac{y_t^{(s,+a)}-y_t^{(s,-a)}}{2a},
 \label{eq:s-finite-response}
\end{equation}
where $a>0$ is the perturbation amplitude and $\ell=t-s$ is the response lag. On a smooth path, this central secant approaches $\partial y_t/\partial p_s$ as $a\rightarrow0$. The measured response can decay while the transported state tangent remains nonzero, because $y_t$ observes only its projection onto the chosen readout.

Let $r$ collect these finite responses over the observation times or readout components being compared, and let $\widehat r$ be the corresponding tangent prediction. Reconstruction is summarized by
\begin{equation}
 \NRMSE=\frac{\|r-\widehat r\|_2}{\|r\|_2},
 \qquad \cos(r,\widehat r)=\frac{r^{\mathsf T}\widehat r}{\|r\|_2\|\widehat r\|_2}.
 \label{eq:s-metrics}
\end{equation}
Here $\|\cdot\|_2$ is the Euclidean norm. NRMSE measures prediction error relative to the observed response norm, while the cosine measures directional alignment independently of overall scale. These are the reconstruction measures used in Figs.~\ref{fig:mathcode}a and \ref{fig:t8-anchor}b. Finite action selection depends on candidate ordering, so the response metrics are accompanied by discrete action agreement and exhaustive-winner recall.

For an independent training history $h$, define the paired decision effect
\begin{equation}
 \Delta_h=\J_h^{\mathrm{comp}}-\J_h^{\mathrm{full}},
 \label{eq:s-history-effect}
\end{equation}
where $\J_h^{\mathrm{full}}$ is the locked terminal objective under full transport and $\J_h^{\mathrm{comp}}$ is the objective under the designated comparator. Positive $\Delta_h$ favors full transport; Fig.~\ref{fig:four-arm}a plots this effect for each comparator. Checkpoints, future tapes, candidate sizes, and control windows within the same history are repeated conditions; the independent history is therefore the inferential unit. Mean-effect intervals resample the history-level values $\{\Delta_h\}$ with the ordinary nonparametric bootstrap~\cite{efron1979}. The exact sign test uses only the number of positive history effects. For 10 positive effects among the 12 four-arm histories, its one-sided probability under equiprobable independent signs is
\begin{equation}
 \Pr\{\operatorname{Binomial}(12,1/2)\ge10\}=\frac{79}{4096}=0.0192871.
 \label{eq:s-sign-test}
\end{equation}
Monte Carlo chains and contiguous token readouts contain local dependence, which is retained by circular-block or contiguous-block resampling~\cite{kunsch1989}.

For candidate setting $g$, let $k_g^*$ be the exact finite-amplitude winner from exhaustive rollout and let $\mathcal S_{g,q}$ contain the neutral action and the $q$ nonneutral candidates retained by a ranking rule. Winner recall at nonneutral rollout budget $q$ is
\begin{equation}
 \operatorname{Recall}(q)=\frac{1}{G}\sum_{g=1}^{G}\mathbf 1\!\left\{k_g^*\in\mathcal S_{g,q}\right\},
 \label{eq:s-winner-recall}
\end{equation}
where $\mathbf 1\{\cdot\}$ is the indicator function and $G$ is the number of evaluated settings. Figure~\ref{fig:cost}b reports this quantity at $q=8$. Candidate-size rows are nested within libraries, multiple libraries share a path, and early and late paths can share a training history. Pooled recall summarizes all 54 Ising and 108 Math--Code settings, while held-path, held-history, and cluster-maximum summaries measure sensitivity to the shared paths and histories. Because these settings are not exchangeable calibration units, the split-conformal framework does not apply directly~\cite{shafer2008conformal,oliveira2024nonexchangeable,jin2023selection}; the reported summaries are empirical calibration diagnostics.

\FloatBarrier
\suppsection{Finite actions, branch compatibility, and exact reranking}{sup:sec-finite-actions}

Section~\ref{sec:response-theory} ranks the seven-action library directly and screens larger libraries before exact reranking. The finite-amplitude remainder, clipping-branch criterion, and shortlist rule are specified below.

\subsection{Finite-amplitude remainder and branch screen}
\label{sup:sub-remainder-screen}

Let $\delta x_t=x_t^{\mathrm{cand}}-x_t^{\mathrm{neutral}}$ be the state displacement between a candidate and the neutral trajectory, and let $\delta p_t$ be their loss-weight difference. Expanding the update map $F_t$ around the neutral path gives
\begin{equation}
 \delta x_{t+1}=A_t\delta x_t+B_t\delta p_t+\rho_t,
 \qquad
 \|\rho_t\|\le\frac{M_t}{2}\bigl(d_t+|\delta p_t|\bigr)^2,
 \label{eq:s-one-step-remainder}
\end{equation}
where $A_t=\partial_xF_t$, $B_t=\partial_pF_t$, $\rho_t$ is the one-step Taylor remainder, $d_t=\|\delta x_t\|$, and $M_t$ is a local Lipschitz constant for the derivative of $F_t$. Vector norms are Euclidean, and $\|A_t\|$ is the induced operator norm. If $\|A_t\|\le\alpha_t$, repeated propagation bounds the terminal state error by
\begin{equation}
 \epsilon_{x,T}\le\sum_{s=t}^{T-1}\left(\prod_{r=s+1}^{T-1}\alpha_r\right)\frac{M_s}{2}(d_s+|\delta p_s|)^2.
 \label{eq:s-state-remainder}
\end{equation}
Here $T$ is the terminal step and $\epsilon_{x,T}$ bounds the difference between the finite state displacement and its tangent prediction. The expression isolates three sources of finite-amplitude error: intervention size, accumulated candidate--neutral separation, and amplification by subsequent Jacobians. Its qualitative dependence guides the choice of small amplitudes and the exact reranking of close candidates.

Gradient clipping makes the update piecewise smooth~\cite{dibernardo2008piecewise}. Each candidate schedule is connected to neutral by the homotopy
\begin{equation}
 \lambda\in\{0,1/8,\ldots,1\},\qquad p_{t+j}(\lambda)=p_0+\lambda a u_j,
 \label{eq:s-homotopy}
\end{equation}
where $j\in\{0,\ldots,H-1\}$ indexes the update within the window and $\lambda$ interpolates from neutral to the full action. A candidate is branch-compatible when every sampled point follows the neutral clipping branch and its gradient norm remains at least 0.002 from the unit clipping threshold. The screen separates sampled changes of clipping regime from smooth curvature within a fixed branch. Response reconstruction measures the latter: the tested SGD condition passes the sampled screen but has a large finite-amplitude reconstruction error.

\subsection{Ranking, exact reranking, and empirical shortlist calibration}
\label{sup:sub-shortlist-calibration}

For candidate schedule $u_k$, let $\J_k(a)$ be the exact terminal objective after executing it at amplitude $a$. Its tangent decomposition is
\begin{equation}
 \J_k(a)=L_k(a)+R_k(a),\qquad L_k(a)=\J_0+a g^{\mathsf T}u_k,
 \label{eq:s-score-remainder}
\end{equation}
where $\J_0$ is the neutral terminal objective, $g=(g_t^{\mathrm{full}},\ldots,g_{t+H-1}^{\mathrm{full}})$ contains the source-time derivatives from one reverse adjoint, $L_k$ is the tangent score, and $R_k$ is its finite-amplitude remainder. Dot products $g^{\mathsf T}u_k$ rank the library before exact rollouts evaluate the retained candidates. If $|R_k|\le\epsilon$ for every candidate, the tangent ordering $L_u<L_v$ is preserved whenever $L_v-L_u>2\epsilon$. Candidates closer than this margin require finite-amplitude comparison.

A narrow fixed shortlist succeeds when tangent-score gaps dominate the candidate-dependent remainders. In the completed development libraries (mixed seeds A and B and the balanced library), $q=8$ gives exact-action agreement of 15/18, 16/18, and 13/18; the diagnostic width $q=16$ gives 17/18, 17/18, and 18/18. Since the second stage evaluates every retained action exactly, these misses occur when the exhaustive winner lies outside the shortlist. The development results motivate a width that adapts to the observed finite-amplitude ranking error.

Let $K$ be the number of nonneutral candidates, $L_{\min}=\min_kL_k$, and $k^*$ the exhaustive finite-amplitude winner on a completed development setting. The normalized tangent-score gap of that winner is
\begin{equation}
 Z=\frac{\max(0,L_{k^*}-L_{\min})}{a^2\sqrt{2\log K}}.
 \label{eq:s-adaptive-margin}
\end{equation}
The one-sided empirical order statistic $c_{0.90}$ of the development values sets the fresh-path tolerance
\begin{equation}
 \epsilon(K,a)=c_{0.90}a^2\sqrt{2\log K},
 \qquad
 \mathcal S(K,a)=\{k:L_k-L_{\min}\le\epsilon(K,a)\}.
 \label{eq:s-adaptive-shortlist}
\end{equation}
Exact rollout then selects the lowest-objective action in $\mathcal S(K,a)$ together with the neutral action. The factor $a^2$ follows the leading Taylor remainder, while $\sqrt{2\log K}$ supplies an empirical candidate-count correction motivated by light-tailed extreme-value scaling. Recalibrated alternatives proportional to $a^2$, $a^2\sqrt{2\log K}$, and $a^2(2\log K)$ give the same outcome on the available fresh rows. The calibration therefore fixes an empirical shortlist rule; the dependence-aware interpretation is given in Supp.~\ref{sup:sec-statistics}.

\begin{figure}[H]
 \centering
 \includegraphics[width=0.94\textwidth]{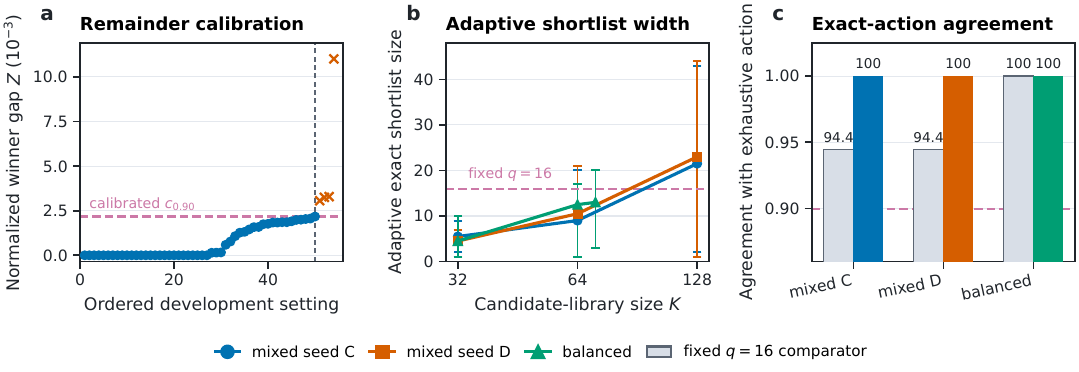}
 \caption{Ising shortlist calibration and fresh-path evaluation. (a) The 50th of 54 ordered development gaps fixes $c_{0.90}$. (b) Applying this coefficient on fresh paths gives the median and range of the adaptive shortlist size for each $K$ and library. (c) Adaptive reranking agrees with the exhaustive action in 18/18 settings per library; fixed $q=16$ misses one $K=128$ action in each mixed library.}
 \label{fig:s-shortlist}
\end{figure}

Fixed $q=16$ already agrees with exhaustive search in 52 of the 54 fresh settings, leaving two large-library actions unresolved. Figure~\ref{fig:s-shortlist} shows what the adaptive calibration adds. The 50th of 54 development gaps fixes $c_{0.90}$ and recalls 50/54 development winners [Fig.~\ref{fig:s-shortlist}(a)]. Applied unchanged to new histories, tapes, and library seeds or orderings, the rule reaches 18/18 agreement in each fresh library [Fig.~\ref{fig:s-shortlist}(c)]. Both improvements occur at $K=128$ on the same early path, where the exhaustive winners have tangent ranks 23 and 22. The adaptive shortlists expand to 43 and 44 candidates on these cases and include both winners [Fig.~\ref{fig:s-shortlist}(b)]. Across all fresh settings, the library-level median widths are 7.5, 7.0, and 9.0. The result is a targeted correction of two large-library ranking failures, obtained by concentrating additional exact rollouts on the affected path.

\FloatBarrier
\suppsection{Additional evidence for response, control, and ranking}{sup:sec-supporting-results}

The response, control, and cost measurements underlying Sec.~\ref{sec:results} and Figs.~\ref{fig:four-arm}--\ref{fig:cost} of the main text are collected here. Each subsection expands one main-text result and relates its numerical diagnostics to the corresponding dynamical or decision-level interpretation.

\subsection{Response reconstruction, optimizer memory, and scale}
\label{sup:sub-response-scale}

The Math--Code full-state tangent reconstructs paired finite responses without a fitted response kernel. At 0.3M, the pathwise NRMSE and cosine similarity are 0.0239 and 0.9997 for the antisymmetric readout $m=(L_A-L_B)/2$, and 0.0490 and 0.9988 for the symmetric readout $e=(L_A+L_B)/2$. The corresponding 1M values are 0.0138/0.9999 and 0.0304/0.9995. Thus, on a specified eight-step path, the local derivative remains accurate at both tested scales.

The optimizer interventions identify how this response is carried. Among the 62 moment-bearing cells that remain on a common smooth clipping branch, full-state transport outperforms moment-tangent deletion in every case. When $\beta_1$ is reduced, the projection of the first-moment response onto its baseline direction falls to 0.4238 of the baseline value. Reducing $\beta_2$ increases the norm of the second-moment response by a factor of 20.70. The common-amplitude SGD comparison behaves differently: its symmetric-response RMS is 11.5 times the AdamW baseline and its NRMSE is 1.46, even though the clipping branch does not change. The fixed perturbation therefore produces a much larger effective displacement under SGD, for which the local linearization is no longer accurate. The finite-amplitude scale of this displacement is distinct from the AdamW coordinate-deletion result.

The accurate local derivative at 1M does not by itself ensure consistent control over a long suffix. Repeated decisions continually separate the controlled and neutral trajectories, changing the subsequent Jacobians about which each window is linearized. Nonlinear remainders can therefore accumulate even when every individual eight-step derivative is accurate, which accounts for the lower history consistency observed in the longer 1M closed-loop run.

\begin{figure}[H]
 \centering
 \includegraphics[width=0.94\textwidth]{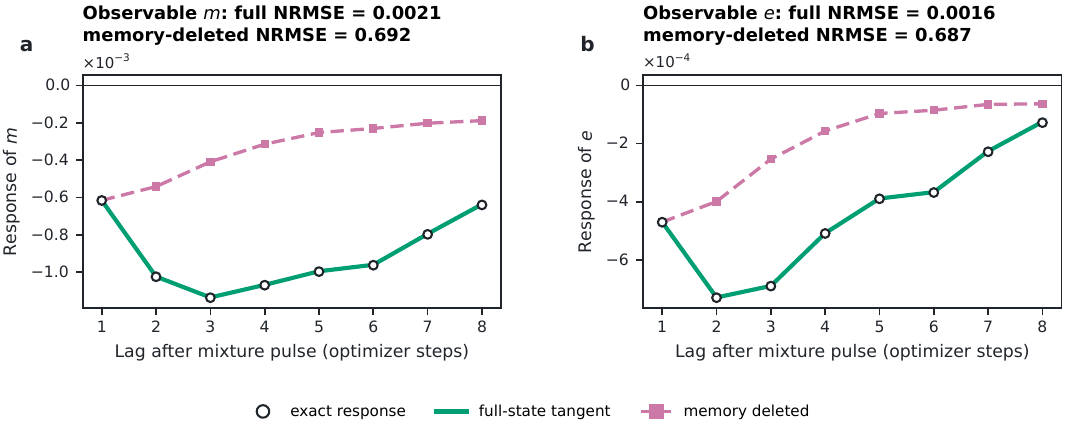}
 \caption{Lag-resolved Ising--CNN response and moment-tangent deletion for (a) the antisymmetric readout $m$ and (b) the symmetric readout $e$. The full-state tangent follows the exact finite response in both panels. Deleting perturbations in the AdamW moment coordinates leaves the first lag unchanged but removes most of the response transmitted to later updates. Curves show path averages; NRMSE is computed pathwise.}
 \label{fig:s-ising-mechanism}
\end{figure}

A direct lag-resolved view comes from the Ising--CNN experiment. Exact finite responses and the full-state tangent remain nearly indistinguishable across all eight lags in Fig.~\ref{fig:s-ising-mechanism}. Moment deletion agrees at the first lag, before retained moment perturbations have propagated through later updates, and then decays too rapidly. Its NRMSE rises from 0.002115 to 0.6923 for $m$ and from 0.001580 to 0.6871 for $e$. The separation after the first update locates the delayed response in the AdamW moment coordinates.

\subsection{Four-arm delayed-transport confirmation}
\label{sup:sub-four-arm}

The prospective four-arm experiment compares full transport with the Adam-aware immediate derivative, neutral training, and VGA on the same 12 histories. The primary full-minus-immediate effects, expressed as reductions in held-out loss, are
\begin{align}
(6.098,9.866,7.395,5.255,6.026,3.418,-2.012,5.935, 8.628,4.800,-1.430,2.522)\times10^{-4}.
\end{align}
Their mean is $4.7082\times10^{-4}$, with 10/12 positive effects and a history-bootstrap 95\% interval of $[2.6596,6.5447]\times10^{-4}$. The full-minus-neutral mean is $7.8052\times10^{-4}$, with 11/12 positive histories and an interval of $[4.9685,10.9430]\times10^{-4}$. Full exceeds VGA in all 12 histories, with a mean difference of $7.7192\times10^{-4}$ and an interval of $[5.1388,11.2400]\times10^{-4}$. The prespecified full--immediate and full--neutral criteria both pass.

The immediate controller also improves over neutral, by $3.0970\times10^{-4}$ on average, and has positive history effects in 9/12 cases. VGA improves by $8.61\times10^{-6}$ and is positive in 4/12. Full and immediate both act in all 96 windows, yet choose the same schedule in only 36. Their different outcomes therefore arise from the ordering of temporal schedules under matched intervention frequency. The full--immediate increment is 60.3\% of the full--neutral benefit; this ratio summarizes the combined effect of delayed propagation, terminal readout, and discrete action selection.

Readout sampling is checked separately. The locked 16-batch-per-domain evaluation gives mean full-minus-immediate benefits of $7.3905\times10^{-4}$ on Math and $2.0259\times10^{-4}$ on Code. A subsequent audit covers 99.984\% of Math and 99.989\% of Code target tokens using disjoint 64-token windows grouped into contiguous 4096-token superblocks. The nearly exhaustive mean benefits are $6.2548\times10^{-4}$ on Math, $2.7318\times10^{-4}$ on Code, and $4.4933\times10^{-4}$ for their equal average; 11/12 combined history effects are positive. A 10,000-replicate contiguous-superblock bootstrap gives $[4.026,4.982]\times10^{-4}$, while joint resampling of histories and superblocks gives $[2.596,6.155]\times10^{-4}$. Agreement with the locked result shows that the measured benefit extends across the test partition.

To distinguish local score changes from state divergence caused by earlier actions, a deterministic replay evaluates both complete candidate-score vectors at every state along the locked full and immediate trajectories. The replay recovers all 96 actions of each controller. Let $i=\arg\min_k S_k^{\mathrm{imm}}$ be the immediate-score winner over branch-compatible candidates, and define its strongest challenger under the full score by
\begin{equation}
 k^*=\arg\min_{k\ne i}\left(S_k^{\mathrm{full}}-S_i^{\mathrm{full}}\right).
\end{equation}
Define
\begin{equation}
 M=S_{k^*}^{\mathrm{imm}}-S_i^{\mathrm{imm}},\qquad
 D=\left(S_{k^*}^{\mathrm{full}}-S_i^{\mathrm{full}}\right)-M.
 \label{eq:s-decision-geometry}
\end{equation}
Here $M$ is the immediate-score margin separating the selected action from the challenger, and $D$ is the change in that relative score produced by delayed transport. With no score ties, the immediate winner changes exactly when $M+D<0$. Ranking and executed-action changes coincide: the scores disagree in 60/96 states along either locked path, and 59 changed history--window locations are shared. At the 12 common initial states, before closed-loop trajectories diverge, the two scores already choose different actions in 6/12 cases.

\begin{figure}[H]
 \centering
 \includegraphics[width=0.96\textwidth]{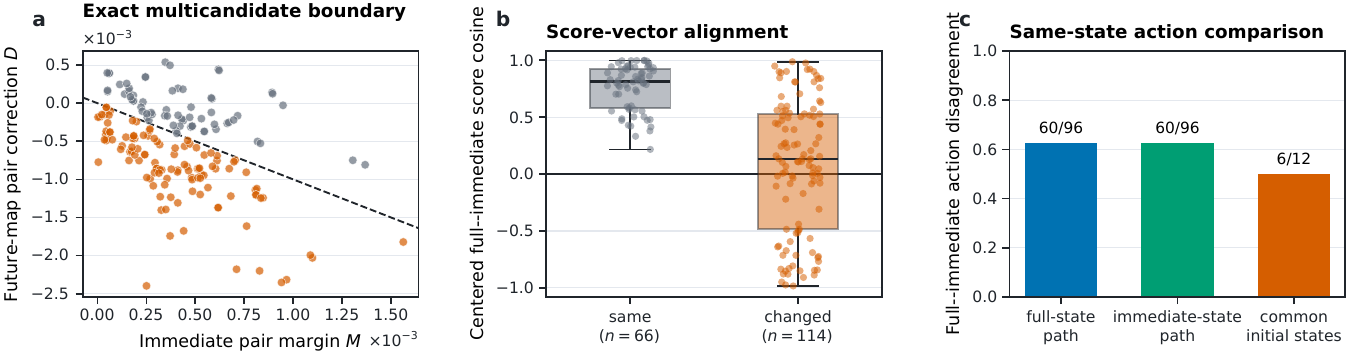}
 \caption{Same-state decision geometry in the four-arm Math--Code experiment. (a) Gray points retain the immediate winner, orange points change it, and the dashed line is the exact multicandidate boundary $D=-M$. (b) Centered full--immediate score cosines for the retained- and changed-winner groups; box lines give medians and quartiles. (c) The two rankings disagree at 60/96 states along each replayed trajectory and at 6/12 common initial states.}
 \label{fig:s-decision-geometry}
\end{figure}

The replayed states separate exactly at $D=-M$ in Fig.~\ref{fig:s-decision-geometry}(a): points below this line receive a future-map correction large enough to overturn the immediate winner. The normalized immediate margins have similar medians when the winner is retained or changed, 1.60 and 1.55, whereas the median centered full--immediate score cosine falls from 0.814 to 0.130 [Fig.~\ref{fig:s-decision-geometry}(b)]. Comparable margins together with the loss of score-vector alignment associate the action changes with a broader reordering of the candidate library. The common-state disagreements in Fig.~\ref{fig:s-decision-geometry}(c) occur before the closed-loop paths separate and isolate the direct contribution of delayed transport to the ranking.

\subsection{Future-path conditioning and anchored Ising control}
\label{sup:sub-future-conditioning}

The crossed audit varies the current checkpoint and the next eight minibatches independently. Centered schedule-score variation decomposes into 10.3\% state--schedule, 17.8\% tape--schedule, and 71.9\% state--tape--schedule interaction. On each realized tape, full actions improve 88/96 cells and all four histories, and beat the immediate comparator in 57/61 informative cells, where ``informative'' denotes a nonzero pairwise benefit difference. Averaging over the other tapes gives a leave-one-tape-out mean benefit of $-1.61\times10^{-5}$ and positive history means in only 1/4 cases. The same optimizer state thus assigns different schedule orderings to different upcoming minibatch sequences, and averaging those orderings can cancel the preference relevant to the realized path.

\begin{figure}[H]
 \centering
 \includegraphics[width=0.94\textwidth]{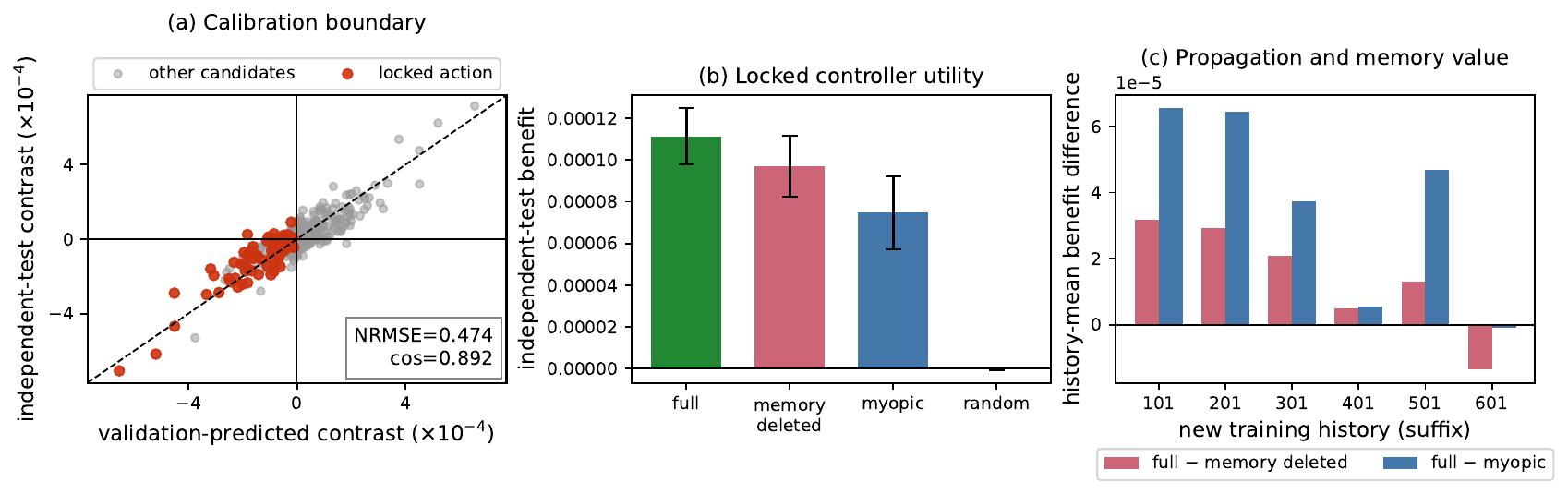}
 \caption{Specified-future-path confirmation. (a) Validation scores preserve useful local action ordering despite imperfect calibration of the full test-contrast vector; red points denote the locked actions. (b) Full transport gives the largest mean independent-test benefit among the four policies. (c) Full exceeds memory deletion and the immediate derivative in five of six history means. Error bars in (b) show standard errors across histories.}
 \label{fig:s-known-tape}
\end{figure}

With the future path fixed, full actions improve 60/72 cells and all six history means. They beat the immediate comparator in 30/41 informative cells and the memory-deleted policy in 22/32, with positive history means in five of six cases for each comparison. Candidate-contrast amplitudes transfer imperfectly from validation to test data [NRMSE 0.474 and cosine 0.892 in Fig.~\ref{fig:s-known-tape}(a)], yet the selected actions have positive test utility [Fig.~\ref{fig:s-known-tape}(b)]. Action selection uses the local ordering near the best candidate, while NRMSE accumulates errors over the entire contrast vector. The history-level differences in Fig.~\ref{fig:s-known-tape}(c) show that delayed propagation and moment transport improve the selected action in five histories; the sixth gives small negative differences for both comparisons.

\phantomsection
\label{sup:sub-binder-control}

The specified-path Ising design also permits a prospective test against the independently constructed Binder anchor. This confirmation retains the fixed Binder artifact while replacing the neural histories, future tapes, calibration chains, and test chains. Full and exhaustive policies agree in 11/12 cells. On the new chains, the mean absolute neural-root error $|\widehat T_N-T_B|$ falls from 0.02435 to 0.01880; all six histories improve, with a block-aware interval of $[0.00258,0.00839]$ for the reduction. Candidate-contrast NRMSE remains above its calibration threshold because it measures the amplitudes of all candidate effects, whereas action agreement depends on the lowest-risk candidate. The more precise Binder estimate combines a direct thermodynamic observable across several lattice sizes; the classifier-dependent $L=16$ crossing $\widehat T_N$ supplies the differentiable training readout controlled here.

\subsection{Equal-budget winner recall, transfer, and cost}
\label{sup:sub-ranking-cost}

For larger libraries, each score retains neutral and selects $q$ nonneutral candidates for exact rollout. At $q=8$, exhaustive-winner recall for full, memory-deleted, immediate, and random ranking is 48/54, 43/54, 20/54, and 0.2959 in Ising, and 108/108, 98/108, 44/108, and 0.1550 in Math--Code. This ordering holds for every tested $q\in\{1,2,4,8,16\}$. On fresh Ising paths, the adaptive shortlist also includes and recovers the exhaustive winner in all 54 settings.

\begin{figure}[H]
 \centering
 \includegraphics[width=0.94\textwidth]{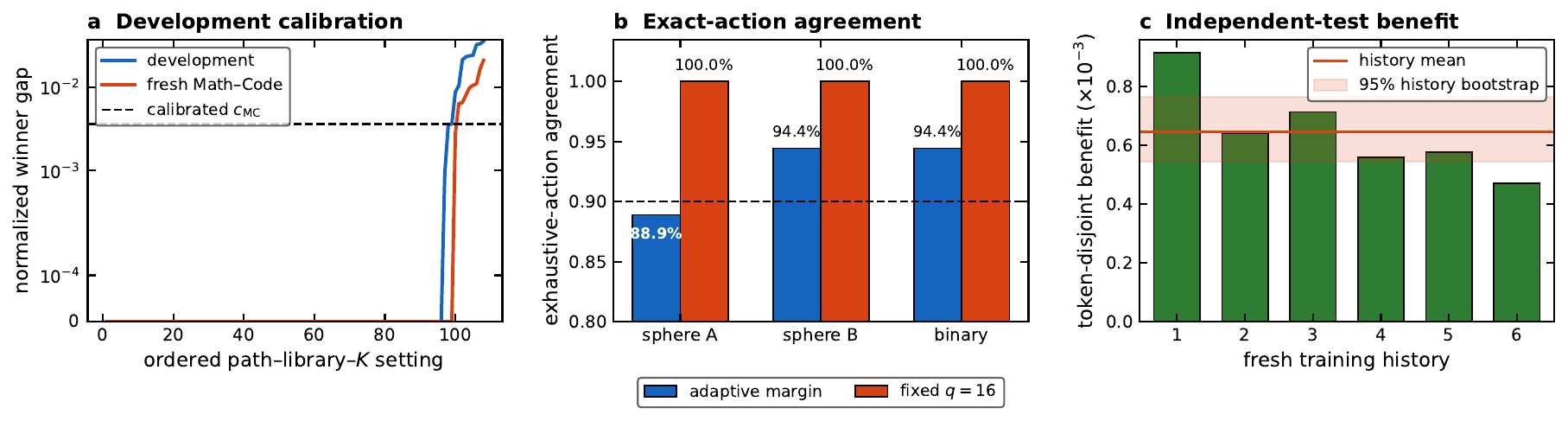}
 \caption{Math--Code transfer of transport-guided exact reranking. (a) Ordered normalized winner gaps for development and fresh settings, together with the frozen Math--Code calibration coefficient. (b) Fixed $q=16$ recovers every exhaustive winner; the adaptive rule gives 88.9\%, 94.4\%, and 94.4\% agreement in the three libraries. (c) Every adaptive action improves token-disjoint test loss; the horizontal line and band show the history mean and its 95\% bootstrap interval.}
 \label{fig:s-mathcode-transfer}
\end{figure}

Three fresh Math--Code libraries test transfer to two random zero-sum spheres and the complete balanced binary library, all with matched aggregate exposure. Fixed $q=16$ recovers all 108 exhaustive winners; the adaptive coefficient recovers 100/108, with 32/36 in sphere A and 34/36 in each remaining library. Their upper-tail gaps extend beyond the development distribution [Fig.~\ref{fig:s-mathcode-transfer}(a,b)], showing that calibrated width depends on local score spacing and library geometry. The locked adaptive actions nevertheless improve token-disjoint loss in all six histories, with mean benefit $6.4612\times10^{-4}$ and a history-bootstrap interval of $[5.4456,7.6545]\times10^{-4}$ [Fig.~\ref{fig:s-mathcode-transfer}(c)].

Winner recall does not include the one-time cost of calibrating the adaptive cutoff. To account for it, one future-path bundle is defined as the three largest libraries at a checkpoint--tape pair, comprising $128+128+71=327$ candidate paths. Ising calibration uses 1962 exact paths and the deployed adaptive rule evaluates 59 paths per bundle on average; Math--Code calibration uses 3924 exact paths and its adaptive rule evaluates six. Charging the complete calibration stage up front gives
\begin{equation}
 N_{\mathrm{break}}^{\mathrm{Ising}}=\frac{1962}{327-59}=7.32,
 \qquad N_{\mathrm{break}}^{\mathrm{MC}}=\frac{3924}{327-6}=12.22
\end{equation}
future-path bundles for the reused library structures. After these points, the saved exact candidate paths, $327-59$ or $327-6$ per bundle, exceed the one-time calibration count. This accounting concerns nonlinear candidate rollouts; the full per-window timing below also includes differentiation, action execution, and readout.

\begin{table}[H]
\caption{Direct 0.3M four-arm CPU cost. Entries are medians of 12 Apple-M4 CPU measurements in seconds; RSS is peak resident memory in GiB from fresh worker processes. ``Screen'' is the branch-compatibility evaluation, ``action'' is the selected eight-step update, and ``readout'' is the held-out evaluation. Component medians need not sum to the median total.}
\label{tab:s-four-arm-cost}
\begin{ruledtabular}
\begin{tabular}{lrrrrrr}
Method & differentiation & screen & action & readout & total & RSS \\
\hline
Ordinary training & 0 & 0 & 0.115 & 0.031 & 0.145 & 0.433 \\
VGA & 0.930 & 5.269 & 0.123 & 0.031 & 6.360 & 0.785 \\
Adam-aware immediate & 1.540 & 5.382 & 0.123 & 0.031 & 7.073 & 0.560 \\
Full, forward & 4.676 & 4.603 & 0.106 & 0.029 & 9.402 & 0.566 \\
Full, adjoint & 0.600 & 4.848 & 0.107 & 0.030 & 5.657 & 1.522 \\
\end{tabular}
\end{ruledtabular}
\end{table}

Forward full, used in the confirmation, takes 1.33 times the total time of the immediate controller because it propagates each source sensitivity through the complete eight-step state map. The adjoint obtains all source-time derivatives in one backward sweep and selects the same action as forward mode on all four timed paths. It takes 0.80 times the immediate total, while retaining more trajectory state and reaching 2.72 times its peak RSS. The two implementations therefore exchange forward-mode time for adjoint memory on this CPU benchmark.

Table~\ref{tab:s-complete-cost} reports the complete per-window costs from the earlier deployment-oriented audit, with $T_{\mathrm{ctrl}}=T_{\mathrm{adjoint}}+T_{\mathrm{exact}}+T_{\mathrm{action}}+T_{\mathrm{readout}}$ and $T_0=T_{\mathrm{train}}+T_{\mathrm{readout}}$.

\begin{table}[H]
\caption{Complete measured Math--Code CPU cost per eight-step window. Times are seconds.}
\label{tab:s-complete-cost}
\begin{ruledtabular}
\begin{tabular}{lrrrrrrr}
Scale & train & readout & adjoint & exact & action & control & $T_{\mathrm{ctrl}}/T_0$ \\
\hline
0.3M & 0.0937 & 0.0275 & 0.4859 & 0.7031 & 0.0945 & 1.3110 & 10.82 \\
1M & 0.1751 & 0.0494 & 0.9201 & 1.2380 & 0.1806 & 2.3881 & 10.64 \\
\end{tabular}
\end{ruledtabular}
\end{table}

The roughly 10.7-fold per-window ratio is dominated by the adjoint and the exact candidate rollouts, which evaluate the specified future map several times before an action is chosen; ordinary training executes that map only once. This ratio measures the complete prospective comparison performed at every window, of which the AdamW update is only one component.

\begin{figure}[H]
 \centering
 \includegraphics[width=0.94\textwidth]{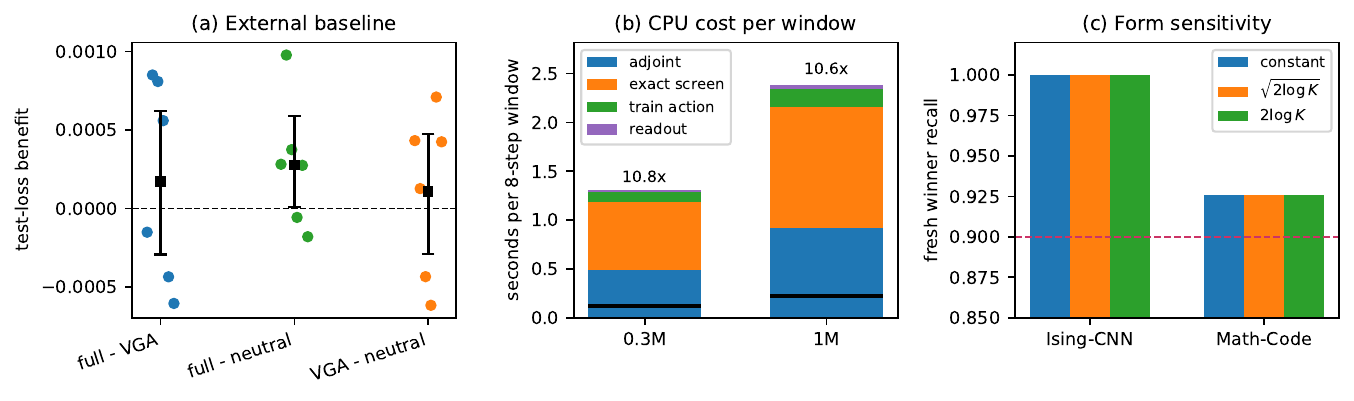}
 \caption{External comparator, complete cost, and candidate-count sensitivity. (a) History-level test-loss benefits in the early full, VGA, and neutral comparison; squares and intervals show means and 95\% history-bootstrap intervals. (b) Measured CPU time per eight-step window, decomposed into adjoint, exact screening, selected-action training, and readout. Labels give the ratio to ordinary training and readout. (c) Fresh-winner recall after separately calibrating three candidate-count factors; the dashed line marks 90\% recall.}
 \label{fig:s-external-cost}
\end{figure}

Three additional measurements clarify the comparator and cost results in Fig.~\ref{fig:s-external-cost}. The early full--VGA difference changes sign across histories, although full transport improves over neutral on average [Fig.~\ref{fig:s-external-cost}(a)]. VGA measures validation-gradient alignment in parameter space, so its difference from full transport combines current AdamW update geometry with subsequent state propagation. The main confirmation uses the Adam-aware immediate derivative to hold the current update fixed. In the complete timing, exact screening and adjoint differentiation dominate the per-window cost at both scales [Fig.~\ref{fig:s-external-cost}(b)]. Separately calibrated constant, $\sqrt{2\log K}$, and $2\log K$ factors give the same fresh-winner recall in each system [Fig.~\ref{fig:s-external-cost}(c)]. For the tested values of $K$ and the correlated library families, recalibration absorbs their numerical differences, leaving the candidate-count dependence unidentified.

\FloatBarrier
\suppsection{Empirical scope and observed boundaries}{sup:sec-boundaries}

The main experiments establish local decisions on specified eight-step paths. Additional tests vary the training scale and duration, the future path and readout, the optimizer, the candidate library, and the evaluation cost. Table~\ref{tab:s-boundaries} groups the resulting boundaries by their observed source.

\begin{table}[htbp]
\caption{Empirical boundaries of the reported results and their observed sources.}
\label{tab:s-boundaries}
\centering
\begingroup
\setlength{\tabcolsep}{3pt}
\renewcommand{\arraystretch}{1.10}
\begin{tabular}{@{}p{0.17\textwidth}p{0.28\textwidth}p{0.25\textwidth}p{0.24\textwidth}@{}}
\toprule
Dimension & Observed boundary & Supported here & Source of boundary \\
\midrule
Scale and duration & 1M: accurate local response; 9/12 positive histories; one path drifts further by 224 steps & One-window reconstruction extends to 1M & Scale and suffix length vary together; repeated actions change later states and Jacobians \\
\midrule
Unknown future & Leave-one-tape-out benefit $-1.61\times10^{-5}$; 1/4 histories positive & Utility on committed future paths & State--tape--schedule interaction contributes 71.9\% of score variation \\
\midrule
Cross-readout & NRMSE/cosine $=0.474/0.892$; calibration gates missed & Locked actions can improve test loss & Action choice uses local ordering; NRMSE weights all contrast amplitudes \\
\midrule
Optimizer family & SGD response is inaccurate at the common amplitude & Moment transport in the tested momentum and AdamW settings & The same pulse gives $11.5\times$ the AdamW symmetric-response RMS \\
\midrule
External comparator & Early full--VGA difference is positive in 3/6 histories & VGA mixes current-update geometry with delayed transport & The VGA score omits AdamW preconditioning and moment-state updates \\
\midrule
Shortlist calibration & Sphere A gives 32/36; recalibrated candidate-count forms tie & Fixed $q=16$ exact reranking gives 108/108 & Winner-gap geometry changes across libraries; recalibration absorbs the tested $K$ factors \\
\midrule
Physical benchmark & Neural-root error decreases but exceeds the Binder error & Control of a physics-anchored neural readout & Binder uses a direct observable at several $L$; the neural root uses one $L=16$ classifier \\
\midrule
Efficiency & Complete CPU control costs $10.6$--$10.8\times$ ordinary training per window & Ranking reduces exact candidate rollouts & Adjoint construction and exact screening repeat the future map before each action \\
\bottomrule
\end{tabular}
\endgroup
\end{table}

The dynamical boundaries arise at different points in the response construction. One-window accuracy at 1M concerns a derivative about the current path; repeated control moves the state and changes the Jacobians used in later windows. Removing the realized future tape changes the drive entering the propagator, and the 71.9\% state--tape--schedule interaction then dominates the action ordering. Changing the readout alters the terminal projection of the transported tangent, so useful local ordering can coexist with inaccurate transfer of the full contrast vector. In the SGD comparison, the same nominal pulse produces a symmetric-response RMS 11.5 times the AdamW value, moving the finite experiment outside the accurately reconstructed amplitude range.

The two Math--Code comparator cohorts answer different questions. In the early six-history experiment, full transport exceeds VGA in 3/6 histories, below the stated history-consistency criterion. VGA measures validation-gradient alignment in parameter-space geometry, whereas the executed update contains AdamW preconditioning and moment state. Its difference from full transport therefore mixes current-update geometry with delayed propagation. The later 12-history experiment uses new histories and an Adam-aware derivative of the same current update, leaving subsequent state transport as the principal distinction. This follow-up protocol was fixed before those new histories were evaluated.

Dependence within a training history also affects shortlist calibration. Leave-one-history-out winner recall is 44/54 for Ising and 97/108 for Math--Code, below the corresponding pooled summaries. For a 90\% empirical order statistic, the index $\lceil(n+1)0.90\rceil$ is 7 when $n=6$, exceeding the number of Ising paths. At $n=12$ for Math--Code, the index is 12, so the cluster-level value is the sample maximum and is substantially larger than the setting-level coefficient. The shortlist threshold is consequently treated as an empirical risk budget, with no finite-sample coverage level assigned.

\FloatBarrier
\suppsection{Reproducibility and data provenance}{sup:sec-reproducibility}

The public repository provides the training and analysis code, configuration files containing the reported input parameters, and the plotting scripts used for the figures. Run identifiers and file hashes bind each result to its training state, future minibatch path, candidate library, selected action, independent readout, and implementation.

Each confirmation fixes the protocol and implementation before unused histories and future tapes are assigned; the resulting checkpoints are hashed. Controller predictions, branch records, actions, and terminal states enter the experiment lock before the independent readout is evaluated. Separate identifiers for histories, checkpoints, tapes, libraries, and readouts keep repeated conditions grouped by training history. Sensitivity analyses have separate records linked to the same lock.

A claim-to-artifact map links each figure and table to its configuration, lock, machine-readable result, and plotting script. Environment records, checkpoint and data manifests, verification tests, and a protocol ledger record the provenance of the released analyses and the inputs used to regenerate the reported summaries.

Late shortlist and timing records use PyTorch 2.12.0; several earlier response records use PyTorch 2.13.0, with the version stored per run. Timing uses float32, deterministic algorithms where supported, four CPU threads, and an Apple M4 CPU. Dynamic CPU frequency makes absolute seconds hardware-specific; component times and ratios support comparisons within the recorded environment. Original code uses the MIT License; upstream licenses govern corpora and checkpoints.

\end{document}